\documentclass[lettersize, journal]{IEEEtran}
\usepackage{amsthm}
\usepackage{amsmath}
\usepackage{amssymb}
\usepackage{graphicx}
\usepackage{indentfirst}
\usepackage{algorithm,algorithmic}
\usepackage[top=1.6cm, bottom=2cm, left=2cm, right=2cm]{geometry}
\usepackage{threeparttable}
\usepackage{multirow}
\usepackage[usenames]{color}
\usepackage{subfigure}
\usepackage{hyperref}

\usepackage{mathtools}

\usepackage[dvipsnames]{xcolor}

\makeatletter
\newcommand*{\rom}[1]{\expandafter\@slowromancap\romannumeral #1@}
\makeatother
\usepackage{xurl}
\usepackage{booktabs}
\usepackage{tikz}

\theoremstyle{definition}

\usepackage{multirow}
\usepackage{array}
\newcommand{\PreserveBackslash}[1]{\let\temp=\\#1\let\\=\temp}
\newcolumntype{C}[1]{>{\PreserveBackslash\centering}p{#1}}
\newcolumntype{R}[1]{>{\PreserveBackslash\raggedleft}p{#1}}
\newcolumntype{L}[1]{>{\PreserveBackslash\raggedright}p{#1}}
\usepackage{amssymb}
\usepackage{pifont}
\newcommand{\cmark}{\ding{51}}%
\newcommand{\xmark}{\ding{55}}%
\usepackage[font=footnotesize]{caption} 
\usepackage{lipsum}
\usepackage{threeparttable}
 \newcommand{\revise}[1]{\textcolor{black}{#1}}

\allowdisplaybreaks
\title{A Survey of Machine Learning-Based Ride-Hailing Planning}

\author{
Dacheng Wen,~\IEEEmembership{Student Member,~IEEE},
Yupeng Li,~\IEEEmembership{Member,~IEEE},
Francis C.M. Lau
\IEEEcompsocitemizethanks{
		\IEEEcompsocthanksitem
            Dacheng Wen is with The University of Hong Kong, Hong Kong (e-mail: wdacheng@connect.hku.hk). Work done while Dacheng Wen was under the supervision of Yupeng Li and Francis C.M. Lau.\protect
            \IEEEcompsocthanksitem Yupeng Li (corresponding author) is with Hong Kong Baptist University, Hong Kong (e-mail: ivanypli@gmail.com).\protect
            \IEEEcompsocthanksitem Francis C.M. Lau is with The University of Hong Kong, Hong Kong (email: fcmlau@cs.hku.hk).
	}
}

\begin{document}
	
	\maketitle

	\begin{abstract}
	\revise{
		\emph{Ride-hailing} is a sustainable transportation paradigm where riders access door-to-door traveling services through a mobile phone application, which has attracted a colossal amount of usage.
		There are two major planning tasks in a ride-hailing system: (1) \emph{matching}, i.e., assigning available vehicles to pick up the riders, and (2) \emph{repositioning}, i.e., proactively relocating vehicles to certain locations to balance the supply and demand of ride-hailing services.
		Recently, many studies of ride-hailing planning that leverage machine learning techniques have emerged.
		In this article, we present a comprehensive overview on latest developments of machine learning-based ride-hailing planning.
		To offer a clear and structured review, 
  we introduce a taxonomy into which we carefully fit the different categories of related works according to
  the types of their planning tasks and solution schemes, which include collective matching, distributed matching, collective repositioning, distributed repositioning, and joint matching and repositioning.
		We further shed light on many real-world datasets and simulators that are indispensable for empirical studies on machine learning-based ride-hailing planning strategies. 
		At last, we propose several promising research directions for this rapidly growing research and practical field.
	}
\end{abstract}
	
	
	\section{Introduction}

\revise{
Ride-hailing, powered by platforms (operators) such as DiDi and Uber, has amassed huge user bases in recent years.\footnote{DiDi: \url{https://www.didiglobal.com};  Uber: \url{https://www.uber.com}.}
As has been reported, DiDi has more than $550$ million registered users while Uber has over $5$ million drivers worldwide \cite{jye19transport, iqbal22uber}.
In {ride-hailing}, the platforms manage intelligently their vehicle resources to fulfil riders' traveling requests.
Compared to the traditional street-hailing mode in which the drivers operate all on their own, ride-hailing is more efficient.
In street-hailing, without any intelligent strategies, an idle driver typically would just pick up the first rider s/he runs into.
But with ride-hailing, the platforms can leverage advanced planning algorithms to manage their fleets to achieve higher efficiency in terms of metrics such as total vehicle miles traveled \cite{ma2017morning}, vehicle capacity utilization rate \cite{cramer2016disruptive}, and rider waiting time \cite{feng2020we, hyland2020operational}.
}
\revise{
	Ride-hailing has greatly reduced the need to own a private vehicle due to its high efficiency in transporting people door-to-door, 
 which consequently would also
lessen traffic congestion, energy consumption, and environmental pollution \cite{kpmg20smart, taiebat2022sharing}.
}

\begin{figure*}[t]
	\centering
	\captionsetup{justification=centering}
	\includegraphics[width=0.75\linewidth]{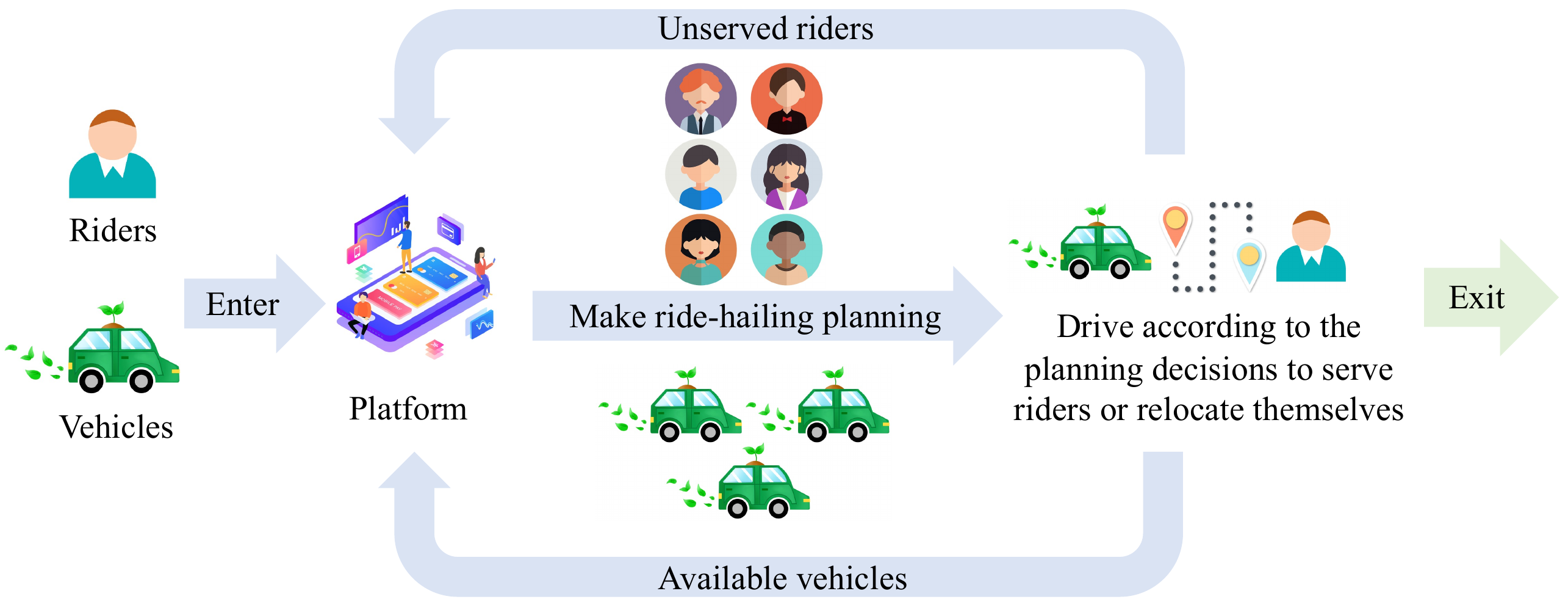}
	\caption{The pipeline of a ride-hailing system.}
	\label{fig:pipeline}
\end{figure*}

A ride-hailing system consists of three types of entities: the platform, drivers, and riders. 
\revise{The operation pipeline of a ride-hailing system and the interaction between the entities are depicted in Fig.~\ref{fig:pipeline}.}
Riders and drivers can enter or exit the ride-hailing system at any time.
\revise{Given the information of riders and drivers, e.g., the origins and destinations of the riders and the real-time locations of the drivers, the ride-hailing platform can make a planning decision accordingly, which includes the \emph{matching} between drivers and riders and the \emph{repositioning} of the vehicles \cite{holler2018deep, holler2019deep, qin2019deep, jin2019coride}. }

\begin{figure}[h]
	\centering
	\subfigure[\revise{Solo-ride}]{
		\includegraphics[width=.4\linewidth]{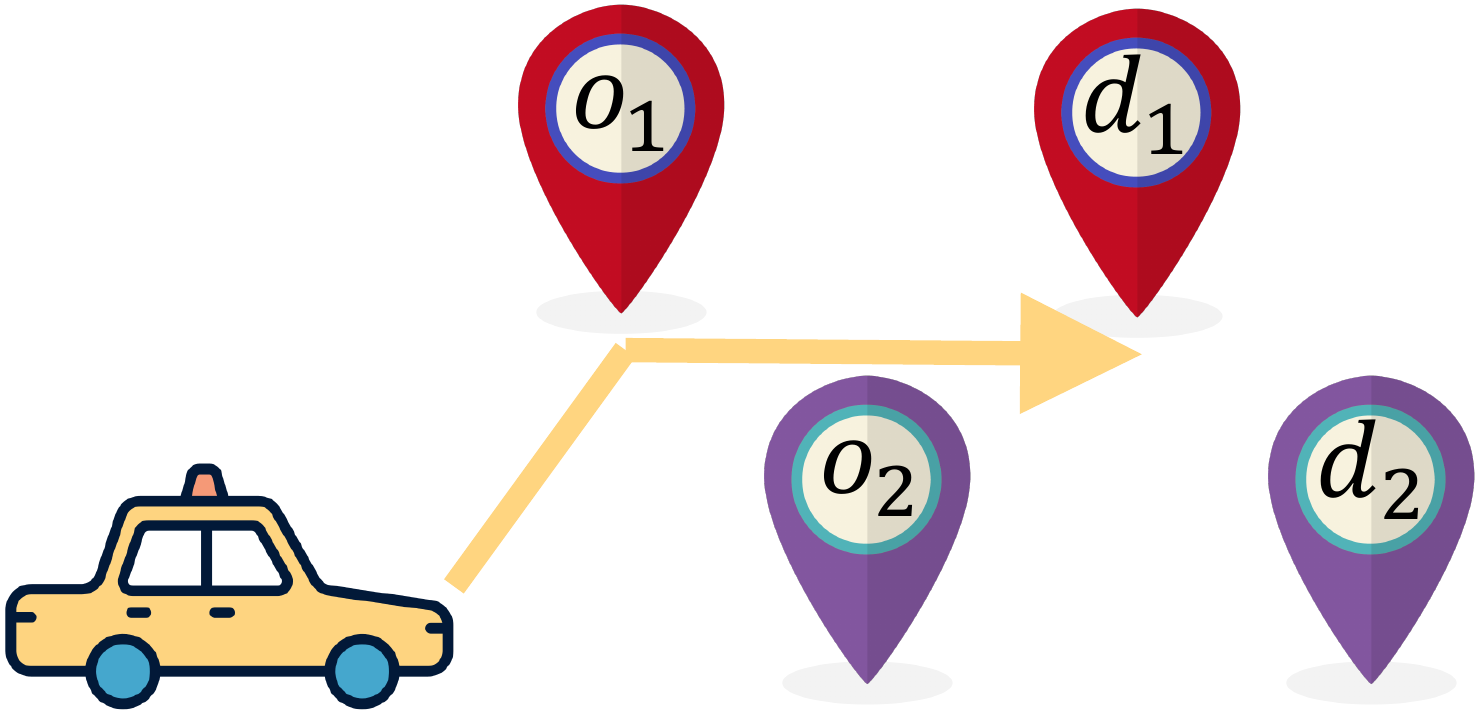}
	}        
	\subfigure[\revise{Shared-ride}]{
		\includegraphics[width=.4\linewidth]{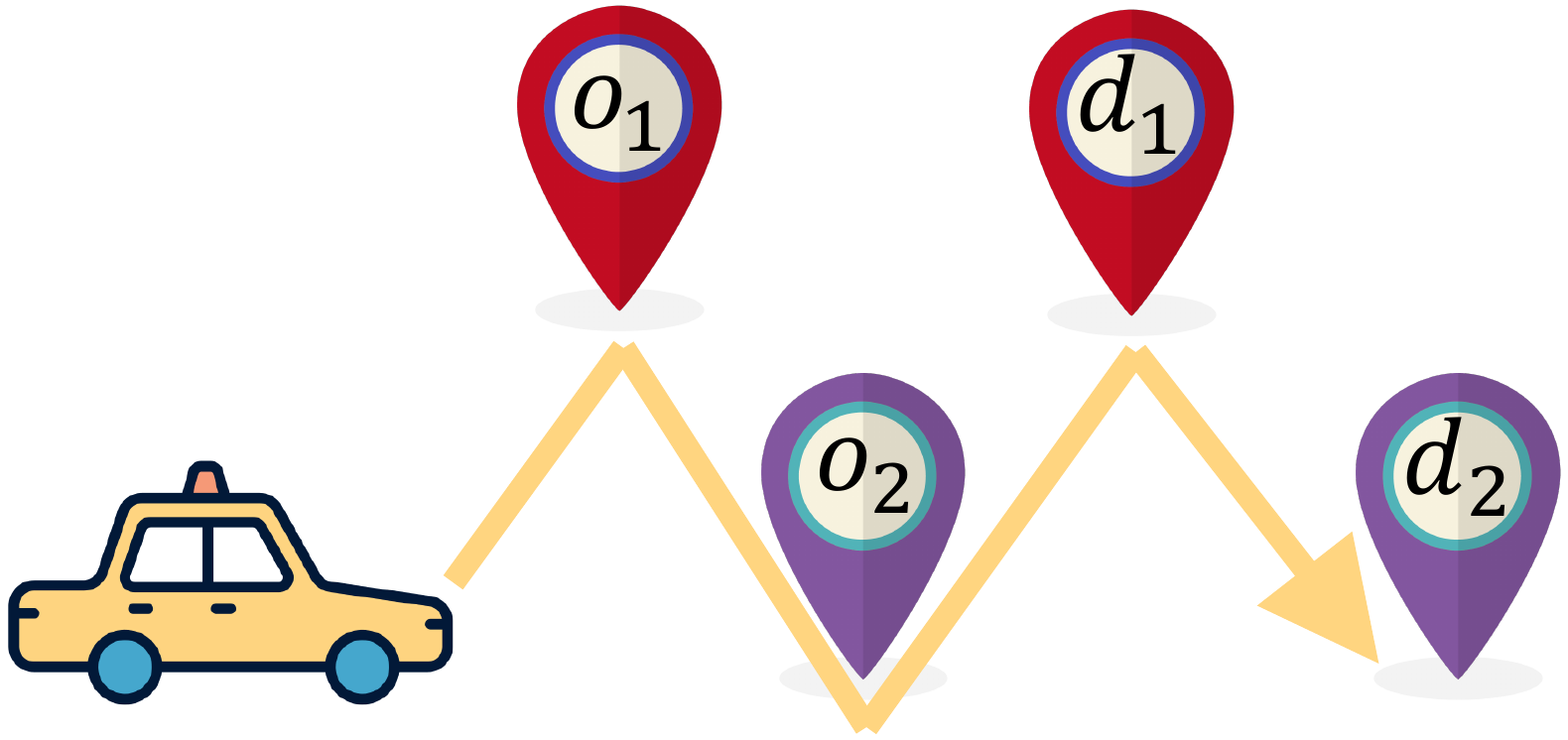}
	}
	\subfigure[\revise{Single-hop}]{
		\includegraphics[width=.4\linewidth]{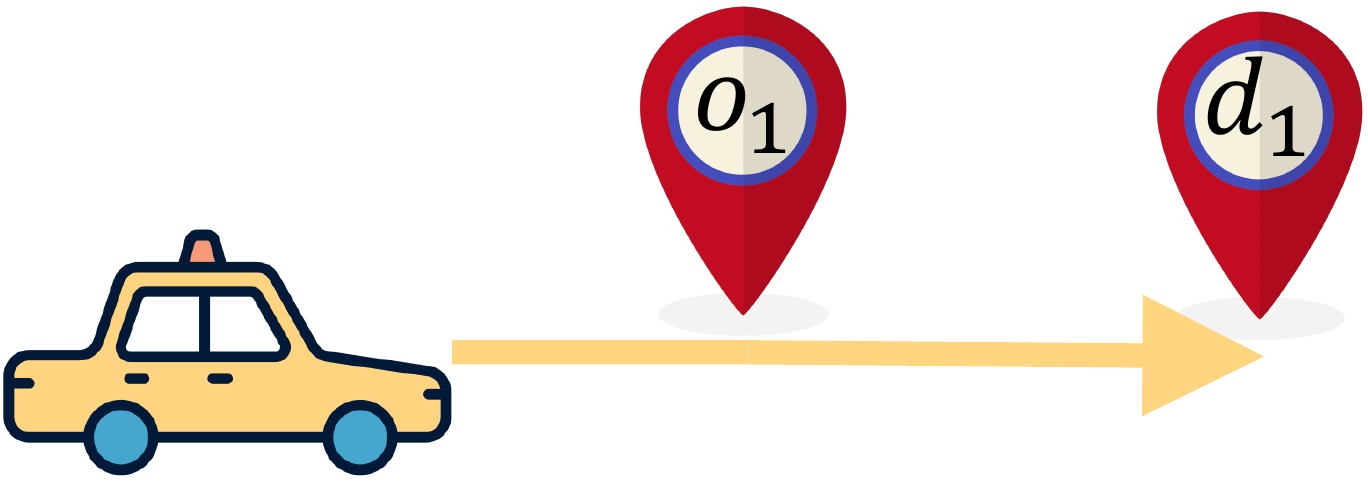}
	}        
	\subfigure[\revise{Multi-hop}]{
		\includegraphics[width=.4\linewidth]{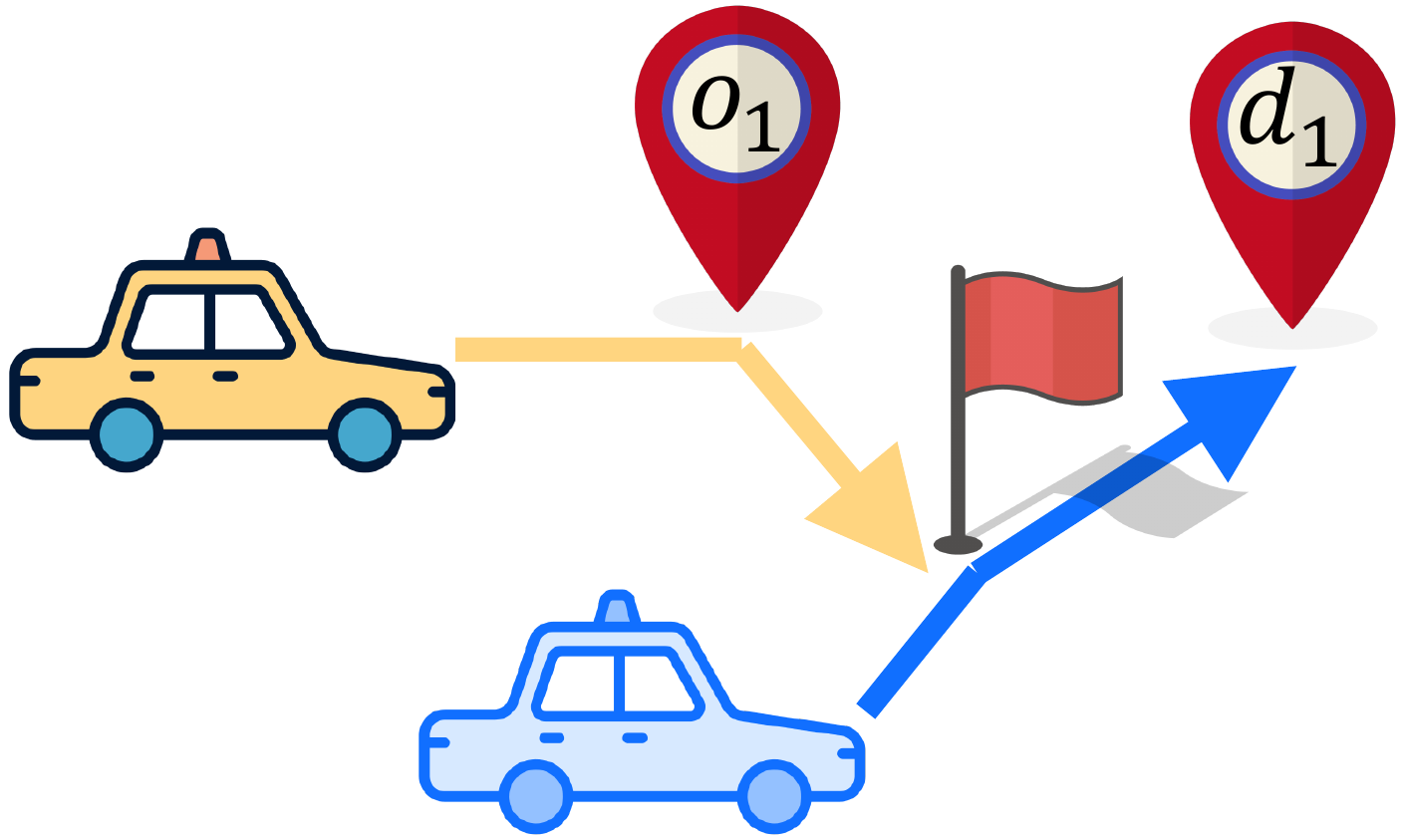}
	}
	\subfigure{
	    \centering
		\includegraphics[width=0.8\linewidth]{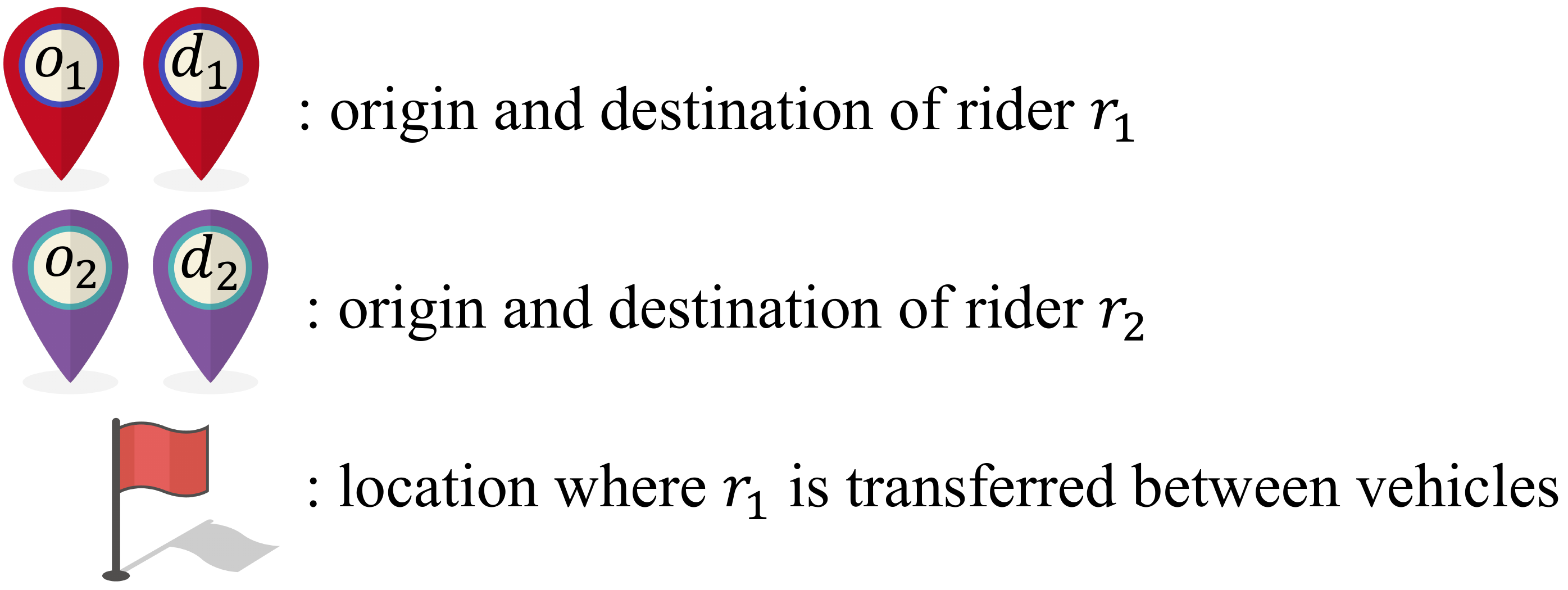}
	}
	\caption{\revise{Solo-ride, shared-ride, single-hop, and multi-hop trips.}}
	\label{fig:matching-type}
\end{figure}

Specifically, matching is to assign available vehicles to riders who have yet to be picked up. 
\revise{Different types of matching are 
offered by existing ride-hailing platforms \cite{tafreshian2020frontiers}.
On the one hand, depending on whether the rider is to travel alone or not, a rider's trip can be a \emph{shared-} or \emph{solo-ride}.
\revise{In the shared-ride mode, multiple riders who share similar routes
can be matched with the same vehicle 
and thus travel together to their destinations.}
In contrast, riders travel alone in solo-ride mode.
\revise{On the other hand, a rider's trip can be \emph{multi-hop} if the rider gets 
transferred from one vehicle to another during the trip; otherwise, it is \emph{single-hop}.}
A visualized explanation of the {shared-ride}, {solo-ride}, multi-hop, and single-hop trips is given in Fig.~\ref{fig:matching-type}.
Accordingly, the matching can be divided into four types: one-to-one (i.e., one hop and solo-ride) \cite{agatz2011dynamic, lloret2017peer, ta2017efficient, long2018ride}, one-to-many 
(i.e., single-hop and shared-ride) \cite{stiglic2016making, regue2016car2work, bei2018algorithms, tamannaei2019carpooling, noruzoliaee2022one}, many-to-one (i.e., multi-hop and solo-ride) \cite{masoud2017real}, and many-to-many (i.e., multi-hop and shared-ride) \cite{agatz2010sustainable, masoud2017decomposition}.}

\begin{figure}[h]
	\centering
	\subfigure[Case 1 (w/o repositioning)]{
		\includegraphics[width=0.98\linewidth]{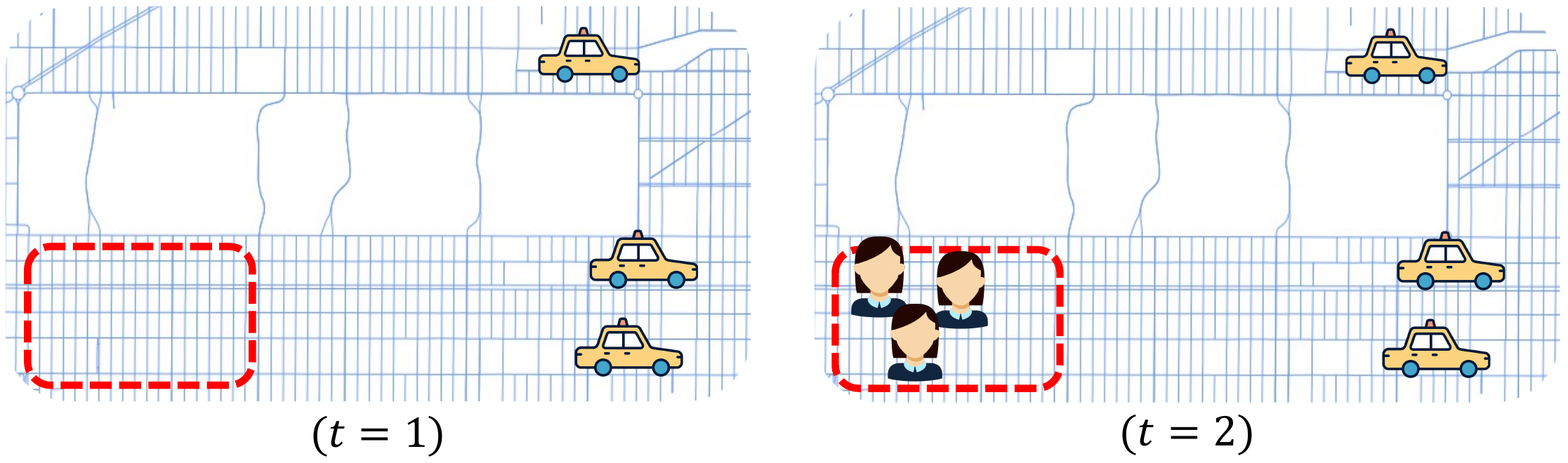}
	}        
	\subfigure[Case 2 (w/ repositioning)]{
		\includegraphics[width=0.98\linewidth]{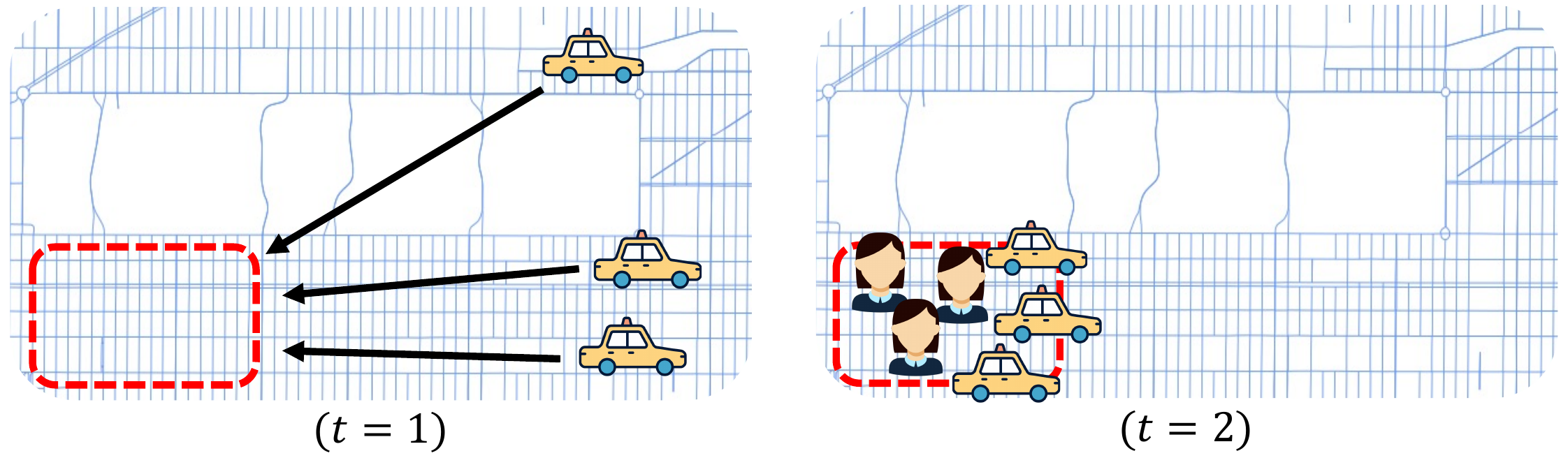}
	}
    \caption{A scenario that consists of three vehicles, three ride requests, and two time steps. In Case 1 (w/o repositioning), the vehicles stay still and wait for new riders at $t=1$.
    \revise{The three ride requests appear in the region circled with red at $t=2$.
    At $t=2$, the riders cannot be served because the vehicles are too far away from the riders.} In contrast, in Case 2 (w/ repositioning), the vehicles relocate to the region at $t=1$ in advance. At $t=2$, the riders can be served.}
	\label{fig:repositioning}
\end{figure}

The purpose of repositioning is to proactively relocate vehicles to certain positions so that these vehicles become better prepared for serving the upcoming riders.
\revise{The repositioning decisions have short-term effects on the driver availability.
In the long run, 
repositioning can substantially change the distribution of drivers across the city.
Thus, repositioning can have a critical impact on how well the potential riders can be matched in terms of certain metrics, e.g., the pickup rate and the waiting time.
\revise{Examples of repositioning are shown in Fig.~\ref{fig:repositioning}.}
We can see from these examples that, with an appropriate repositioning strategy, more riders can be served than if drivers simply stay put while waiting for new riders.}

Planning is the most important task for a ride-hailing platform \cite{holler2018deep, holler2019deep, qin2019deep}.
However, it is a computationally intractable task \cite{alonso2017demand}.
\revise{Specifically, when the size of the planning problem is non-trivial, i.e., the numbers of drivers and riders respectively are large, exact optimization algorithms (e.g., those based on branch-and-cut \cite{liu2015branch}, branch-and-price \cite{qu2015branch, parragh2015dial}, column generation \cite{parragh2012models}, dynamic programming \cite{hame2015maximum}, and Benders decomposition \cite{cortes2010pickup}) all turn out to be infeasible due to their intolerably long running time \cite{ordonez2017dynamic, molenbruch2017typology}.
For example, even when dealing with a small instance that consists of two drivers, six riders, and a city of $1000 \times 1000$ squared distance units, a solution based on the traditional branch-and-bound method takes more than thirty minutes of computation time on an average machine 
\cite{cortes2010pickup}, which is unacceptable for the platforms, drivers, or riders in practice.
Instead, to achieve near-optimal performance of ride-hailing planning in real-time, approximate solutions that can solve problems of real-world sizes, e.g., those using heuristics \cite{luo2011online, chassaing2016els, ritzinger2016dynamic, dutta2018hashing, SHARIFAZADEH2022128, de2022influence, zhou2022scalable, peng2022investigating} and  machine learning (ML) \cite{nguyen2018policy, al2019deeppool, shi2021learning, oda2021equilibrium, haliem2021distributed}, have been proposed.}

\revise{Several literature reviews of ride-hailing planning have been conducted over the past decade.
Some of them focus on matching between drivers and riders, and propose exact or heuristic optimization solutions \cite{agatz2012optimization, furuhata2013ridesharing, mourad2019survey, yan2020dynamic, tafreshian2020frontiers, martins2021optimizing}.
\revise{However, these solutions skip over the vehicle repositioning problem which is a critical component in a ride-hailing system.}
Besides, the superiority of ML techniques in dealing with complex planning tasks has led to an increased number of ML-based methods for ride-hailing planning (e.g., \cite{holler2018deep, lin2018efficient, tang2019deep, holler2019deep, qin2019deep}), which are not covered in the reviews mentioned above. 
\revise{There exist several tutorials and surveys that concentrate on
ML-based transportation \cite{veres2019deep, farazi2020deep, haghighat2020applications, chakraborty2020review}, but they cover only a limited subset of the literature on
ride-hailing planning.}
Qin et al.~\cite{qin2021reinforcement, qin2022reinforcement} focus on a single specific branch of ML, reinforcement learning (RL), in the context of ride-hailing.
Our survey presents a more comprehensive overview of the literature on ML-based ride-hailing planning.}
To sum up, our work has the following contributions:

\begin{enumerate}
	
	\item \emph{New taxonomy.} 
	\revise{To set the stage for a clear review, we create a 
	taxonomy for the literature of ML-based ride-hailing planning. 
 First, we categorize the literature into groups according to their tasks of the planning, including matching, repositioning, and joint matching and repositioning.
Second, we divide the works in each group based on their solution scheme into two divisions, including \emph{collective planning} and \emph{distributed planning}. 
	In the collective scheme, the planning decisions of all drivers and riders are determined jointly. }
	\revise{Whereas in the distributed scheme, each vehicle makes its own decisions of planning without coordinating with other vehicles.
	Existing methods of ride-hailing planning are either collective or distributed.}
	\revise{By treating collective and distributed plannings separately, we can understand better their different decision making processes and the corresponding advantages.
 At last, in each division, we further classify the works based on their different ML techniques and foci and discuss them in detail accordingly.}
	
	\item \emph{Comprehensive review.} 
	\revise{As mentioned above, existing surveys did not include a comprehensive coverage of the works on ML-based ride-hailing planning.}
	\revise{We fill this gap by providing an extensive overview on 
 the latest developments of this topic. }
	
	\item \emph{Abundant resources for empirical studies.} 
	\revise{We examine many public benchmark datasets for ride-hailing planning that are critical for developing and evaluating data-driven ML-based solutions.
	Tools that can be used to simulate the ride-hailing process are important for performance evaluation.
	Thus, we also review the relevant open-source simulators in this survey.}
	
	\item \emph{Future directions.} 
	\revise{We suggest several promising research problems as future directions, which include drivers' and riders' heterogeneity, malicious behaviors, and fairness in ML-based ride-hailing planning.}
	Besides, emerging ML techniques that have been overlooked in ride-hailing planning are discussed.
\end{enumerate}

The rest of this survey is organized as follows.
\revise{The background of ride-hailing planning is given in Sec.~\ref{sec:background}.
Sec.~\ref{sec:review} presents a detailed survey of the literature.
Resources useful for empirical studies of ride-hailing planning are reviewed in Sec.~\ref{sec:resource}.}
Finally, we point out some future directions and draw our conclusion in Sec.~\ref{sec:future} and Sec.~\ref{sec:conclusion}, respectively.

	\section{Background} 
\label{sec:background}
    In this part, we start with the main components of the ride-hailing system and their interactions in Sec.~\ref{sec:background-system}. 
    Then we discuss the ride-hailing problems and their challenges in Sec.~\ref{sec:background-planning}.
    \revise{Our proposed taxonomy for the solution scheme of the methods of existing ML-based ride-hailing planning is introduced in Sec.~\ref{sec:background-scheme}.}

\subsection{\revise{Ride-Hailing System}}
\label{sec:background-system}
    \revise{There are three types of entities in a ride-hailing system: the ride-hailing platform, drivers, and riders.}
    A ride-hailing platform operates in the system as a transportation service organizer.
    It establishes the connections and communications between service providers (i.e., drivers) and service consumers (i.e., riders).\footnote{In this paper, we regard drivers 
    and riders 
    as the supply and the demand, respectively.}
    \revise{To illustrate the interactions among the three types of entities, we give a visualization of the ride-hailing process in Fig.~\ref{fig:pipeline}.}
    At any time, riders and drivers can enter the system by making ride requests and ride offers, respectively, through the ride-hailing platform.
    Upon receiving ride offers and requests, the platform makes \emph{planning} decisions for the drivers and riders.
    A planning decision includes the matching between ride offers and ride requests and the repositioning of vehicles.\footnote{\revise{Similar to the planning, \emph{pricing} is another important decision-making module in the ride-hailing system. Our survey focuses on the ML-based ride-hailing planning where most existing works investigate matching and/or repositioning. Notably, many worldwide leading ride-hailing
platforms such as DiDi decouple the decision-making process of pricing from the ones of matching and repositioning.}}

\subsection{\revise{Ride-Hailing Planning}}
\label{sec:background-planning}
    \revise{We dig into the problems of matching and repositioning as they show up in the ride-hailing planning in Sec.~\ref{sec:background-planning-matching} and Sec.~\ref{sec:background-planning-repositioning}, respectively.
    For each part, the general forms of the problem with problem inputs, decisions, and objectives are presented, followed by the challenges we may face when designing solutions.}
    
    \subsubsection{\revise{Matching and Its Challenges}}
    \label{sec:background-planning-matching}
    Matching is the problem of assigning ride offers to ride requests \cite{alonso2017demand, dickerson2018allocation, xu2018large}.
    \revise{A ride-hailing platform makes matching decisions with the effect of keeping the ride-hailing system function properly.}
    The inputs of the matching problem include the information of all vehicles and riders.
    \revise{Each rider has an origin and a destination.
   	A vehicle has its real-time location, on-board rider(s), remaining capacity, and a predefined destination of the driver, if applicable (depending on whether the driver has her/his own trip to make).
    Besides the vehicles and riders, other information may also be useful, such as supply and demand patterns that are learned from the historical data (e.g., \cite{alonso2017predictive, liu2019globally, lin2019probabilistic, wang2019data}) and estimated time of traveling along different routes (e.g., \cite{jindal2018optimizing, xu2018large, zhou2019multi, liu2022personalized}).
    \revise{Given all the information mentioned above, the platform can make matching decisions on the assignments of ride offers to ride requests.}
    In the mode of shared-rides, multiple riders w.r.t.~different ride requests can travel together in a vehicle to their destinations, subject to some preset constraints, e.g., the maximum detour distance and latest time of arrival of the riders (see \cite{mourad2019survey} for more constraints in shared-rides).
    In solo-ride mode, riders are served alone in vehicles during their entire trips.
    \revise{If multi-hop riding is allowed, a rider can be matched with multiple vehicles that cover
    different hops of her/his way to the destination.}
    \cite{masoud2017decomposition, shah2020sride, xu2020highly}.
    There is a variety of objectives for the matching decisions as revealed
    in the literature.}
    \revise{Mourad et al.~\cite{mourad2019survey} divide the objectives into two categories, operational objectives and quality-related objectives.} 
    The former category is concerned with system-wide metrics, e.g., total vehicle miles traveled and the number of served requests.
    The latter focuses on the quality of ride-hailing services from the perspective of individuals, which is usually measured by riders' waiting time, riders' detour time, drivers' incomes, etc.
    
    \revise{The problem of ride-hailing matching has been proved to be NP-hard
    even in the offline setting where full knowledge of ride requests and ride offers are known in advance \cite{bei2018algorithms, alonso2017demand}.
    The online nature of ride offers and ride requests makes matching even more challenging since future information is not known or uncertain when decisions are made.
    A matching decision at any time can influence the upcoming ride requests and vehicles.
    A \emph{myopic} strategy, such as those in \cite{alonso2017demand, bei2018algorithms, li2020trip}, that matches drivers and riders with an aim of optimizing the outcomes only at
    the current time-step may not satisfy long-term objectives.
    Besides, the scale of the problem instance is usually overwhelmingly large for many metropolitan areas, and thus efficient solutions are very much sought after 
    \cite{agatz2012optimization}.}
    In reality, the uncertainty in the environment, e.g., travel time and traffic condition, could also have an impact on the performance of the matching algorithms. 
    
    \subsubsection{\revise{Repositioning and Its Challenges}}
    \label{sec:background-planning-repositioning}
    \revise{Repositioning is the problem of proactively relocating vehicles to certain positions in advance to prepare them for achieving better outcomes in future matching \cite{lin2018efficient, xu2020recommender, chaudhari2020learn}.
    Though there are drivers who are willing to provide ride offers, vehicle resources can still be underutilized, as ride requests can easily end up unserved without an appropriate repositioning strategy.}
    By properly allocating the vehicle resources, repositioning can lead to high utility of vehicles and high fulfillment of riders \cite{chaudhari2020learn}.
    \revise{To solve the problem of repositioning, the input can be current spatial distributions of vehicles and riders.}
    \revise{Additional inputs could include predictions about future supplies and demands \cite{gao2020learning, zhang2020multiple, guo2021multi, zhang2021mlrnn, ke2021joint, chen2022h}.}
    \revise{The output decisions are the positions to relocate to for all idle vehicles.}
    \revise{Examples of commonly adopted 
    repositioning objectives in the literature include minimization of the idle time of vehicles \cite{o2021using}, the waiting time of riders \cite{ji2020spatio}, the number of unfulfilled ride requests \cite{liu2019globally}, the overall difference between the supply and demand of all regions being considered \cite{guo2021multi}, etc. 
    There are also some 
    objectives that aim to maximize metrics such as gross merchandise volume of the platform \cite{lin2018efficient} and the cumulative driver incomes \cite{jin2019coride}.}
    
    The challenges of the repositioning problem are threefold.
    \revise{Firstly, similar to the matching problem, a ride-hailing platform typically needs to deal with a large number of vehicles in real time \cite{oda2018distributed}.}
    Simple strategies can lead to poor performance.
    \revise{For example, a method that greedily relocates available vehicles to the places that have a concentration of ride requests might leave other places with insufficient supply of vehicles. }
    Making a satisfactory repositioning decision efficiently is imperative.
    \revise{Secondly, the supply and demand themselves are not only changing dynamically over time but also subject to the impacts caused by the repositioning decisions made at the previous time steps, the relation between which, however, is difficult to be modeled in the decision making process \cite{liu2020context}.
    Thirdly, in practice, drivers can, due to self-interest, deviate from the repositioning decisions given by the platform in order to maximize their own profits \cite{oda2018movi, sadeghiyengejeh2021re}. }
    The inability of complete control over the behaviors of vehicle resources can introduce more uncertainty in planning.
    
\subsection{\revise{Schemes of Ride-Hailing Planning}}
\label{sec:background-scheme}
    \revise{In this part, we introduce a 
    taxonomy with which we can divide the literature of ride-hailing planning into two groups based on their solution schemes, namely \emph{collective} planning and \emph{distributed} planning. }
    
   \revise{In collective planning strategies, the matching or repositioning decisions of all drivers are determined jointly.
    The decisions of individuals depend on each other.
    In this way, coordination among the involved drivers can be considered explicitly.
    It helps optimize the decisions globally and hence benefits the performance regarding certain system-wide objectives, such as the maximization of platforms’ incomes and the overall satisfaction of riders’ requirements \cite{he2019spatio}.
    Thus, collective planning is preferable when aiming at the global optimum against the system-wide objectives.}
    
    \revise{As for the distributed scheme, the planning processes of drivers are generally independent of each other.
    Drivers' actions can be decided asynchronously, i.e., in a per driver manner, without any explicit coordination between one another.
    In other words, while planning the decision for a driver, the actions of others are not considered \cite{oda2018movi, al2019deeppool, singh2021distributed, haliem2021distributed}.
    Though it may not be able to achieve the global optimum of certain system-wide parameters as mentioned above in the distributed scheme, it has some advantages in the following aspects over the collective scheme \cite{al2019deeppool, oda2018movi, chau2020decentralized}.
    Firstly, the planning decisions in the distributed scheme can respect each individual's interests, while the collectively planned decisions that aim at the system-wide objectives may not coincide with each driver's or each rider's intentions.
    For example, some drivers might be required to pick up some riders far away from them or relocate to some places with low demand, which are not ideal for the drivers.
    In this regard, drivers may prefer distributed planning to the collective one as they are driven by self interest. 
    Secondly, compared to the collective design, making decisions from the perspective of a single driver reduces the computational complexity in planning. }
    That is, the distributed planning scheme ensures scalability as a result of less coordination between vehicles. 
    This is especially advantageous when dealing with large problem instances in practice, e.g., during peak hours or in crowded cities.
    \revise{Thirdly, higher planning efficiency can reduce the challenges caused by the uncertainty
    of rapidly changing environments, e.g., varying patterns of supply and demand, travel time, and traffic conditions 
    due to
    the fact that distributed planning process takes less time than the collective one and can be done within shorter time slots.} 

\section{Review of ML-based ride-hailing planning}
\label{sec:review}
\begin{figure*}[h]
	\centering
	\includegraphics[width=0.8\linewidth]{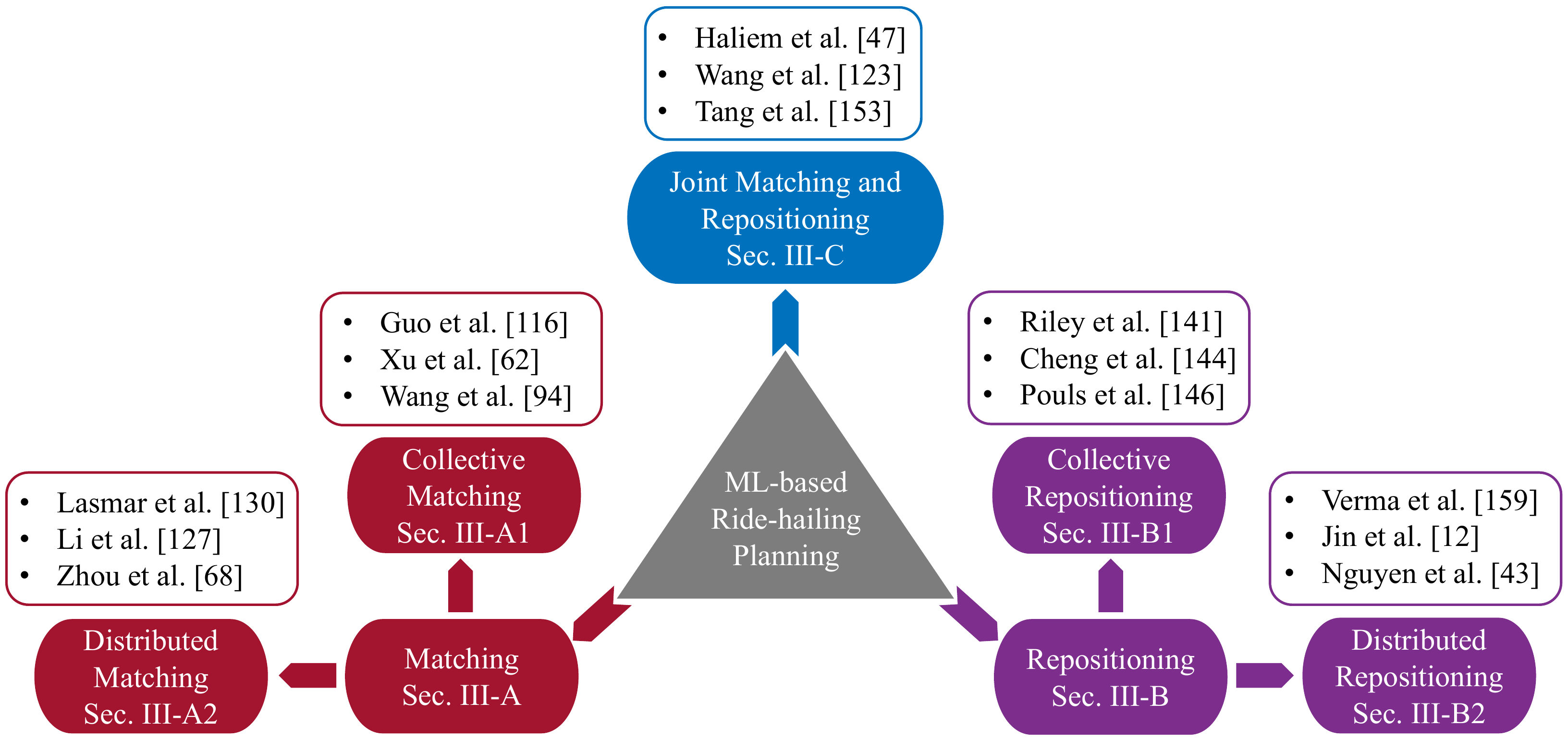}
	\caption{\revise{A taxonomy of the ride-hailing planning literature. 
	In each category, we discuss three works as representative examples.}}
	\label{fig:review-outline}
\end{figure*}

\revise{In this section, we review matching, repositioning, and joint matching and repositioning in Sec.~\ref{sec:review-matching} Sec.~\ref{sec:review-repositioning}, and Sec.~\ref{sec:review-joint}, respectively.}
In each part, we discuss the collective and the distributed strategy separately.
Fig.~\ref{fig:review-outline} gives an outline of the review.

\subsection{Matching}
\label{sec:review-matching}

\subsubsection{Collective Matching}
\revise{RL is a promising technique for solving the matching problem.
Chen et al.~\cite{chen2020order} propose an RL-based solution in which
a deep evaluation network, which is a plain feed-forward neural network, is used to calculate a score for each pair of driver and rider based on the predicted detour distance, vehicle's seat utilization rate, and profit achieved if they get matched.
For each new ride request, the vehicle with the highest score will be assigned to serve the rider.
When the trip of the ride request is finished, the observed reward, i.e., the sum of the increased profit of the driver and, if any, the reduced cost of the rider through sharing the ride with others, is used to guide the learning process of the deep evaluation network.
Agussurja et al.~\cite{agussurja2019state} formulate the matching problem as a two-stage planning process.
In the first stage, ride requests to be scheduled are selected from all the unserved ones, the problem of which is modeled as a Markov Decision Process.
An approximated value iteration algorithm is used to learn the value function for the matching actions.
In the second stage, the final matching decision is made between the selected ride requests and all vehicles based on the learned value function.
\revise{Kullman et al.~\cite{kullman2022dynamic} apply deep RL to develop matching policies whose decisions leverage the Q-value approximations learned by deep neural networks.}
Multi-hop ride-hailing can improve the efficiency of a ride-hailing system.
To find the transfer points for each transferring trip in the multi-hop ride-hailing service, Xu et al.~\cite{xu2020highly} use a multi-layer feed-forward network to predict the reachable areas of vehicles, based on which the search space of possible vehicle pairs and transfer points for transferring riders is pruned.
In this way, the transfer points searching process can be more efficient.
	Wang et al.~\cite{wang2023optimization} also consider the scenario where riders are allowed to transfer between vehicles.
	They leverage RL to learn a policy that estimates the values of all the vehicles, which are then used to compute the optimal matching decisions by integer-linear programming.
The lengths of the time-intervals between the matching decisions can have critical impact in the matching outcomes.
	Specifically, the efficiency of matching may be improved substantially if the matching is delayed by adaptively adjusting the matching time-intervals according to the real-time situation of the riders and drivers.
	Wang et al.~\cite{wang2019adaptive} find that, if riders are willing to wait for a certain amount of time even if there are available vehicles that can serve them right away, the ride-hailing system can achieve better results, for example, in terms of the total vehicle miles traveled.
	In their solution, 
	 they propose to use an RL policy to decide for each rider, at each time step, whether to conduct matching for her/him, or 
	wait for the next time step. 
	Similarly, Qin et al.~\cite{qin2021optimizing} leverage RL in solving the ride-hailing matching problem with dynamic matching time-intervals.}

\revise{Clustering techniques are frequently used in ride-hailing planning.
Hong et al.~\cite{hong2017commuter} propose to use a density-based clustering algorithm, specifically DBSCAN \cite{parimala2011survey}, to identify riders that share similar itineraries based on their historical traveling trajectories. 
To alleviate the computational overhead caused by the large number of distance queries in the matching process, Zhang et al.~\cite{zheng2018order} propose a new clustering algorithm that groups the geographical locations in the road network into different clusters.
Then, the distance between any two nodes is approximated by the distance between the centers of the clusters they belong to. 
Shen et al.~\cite{shen2019roo} propose a spatial-temporal distance metric that measures the similarity of each pair of ride requests.
The ride requests are grouped by a clustering process based on the proposed distance metric.
Then, shared-rides are computed within each group of ride requests.
Another clustering algorithm is proposed by \cite{trasarti2011mining} to extract the mobility profiles from riders' and drivers' historical itineraries.
The matching between riders and drivers is determined based on the similarities between their profiles.}

\revise{An increasing number of collective matching solutions leverage various other ML techniques in planning.
Most of them take social factors of drivers and riders into consideration \cite{mitropoulos2021systematic}.
To mitigate the social barriers in the ride-hailing process, especially in shared-rides, Yatnalkar et al.~\cite{yatnalkar2020enhanced} and Narman et al.~\cite{narman2021enhanced} use Support Vector Machine (SVM) to predict the user social types, e.g., chatty, safety, or punctuality, based on their registered user characteristics.
Riders with similar social characteristics 
would be more willing to share a trip.
Levinger et al.~\cite{levinger2020human} use a feed-forward neural network to predict rider satisfaction levels according to their profile and trip information.
They proposed a stochastic algorithm to compute the matching decision with rider satisfaction level maximization as the objective.
Montazery and Wilson \cite{montazery2016learning, montazery2018new} propose to take into account the user preference in evaluating the weight (benefit) of the matching between each pair of rider and driver, which is given by their proposed support vector machine-based score function.
With the value calculated, the final matching can be obtained by solving an optimization problem in which the sum of the weights of those matched pairs is maximized.
Tang et al.~\cite{tang2020efficient} model various types of information (e.g., driver, rider, travel time, and activity) and their relationships within a ride-hailing system using a Heterogeneous Information Network (HIN) \cite{sun2012mining}.
Each driver or rider is projected to a multi-dimensional embedding (vector) using the skip-gram model \cite{mikolov2013efficient}.
Moreover, the skip-gram is conducted on node sequences obtained by meta path-based random walks originating from the corresponding node within the HIN \cite{dong2017metapath2vec}.
The cosine similarity between the embeddings of each driver-rider pair is then used to identify possible matching.
Zhang et al.~\cite{zhang2017taxi} consider a scenario where each rider is assigned to multiple drivers (to improve the order answer rate), and riders are free from having to enter the details of destinations (to improve the user experience).
They first leverage historical data to model the probability distribution of destinations of each rider based on his/her departure time and location with Bayesian rules, which is followed by predicting the acceptance probability between the rider and available drivers with logistic regression \cite{friedman2001elements} and gradient boosted decision tree \cite{mason1999boosting}.
They propose a hill climbing-based algorithm to solve the matching problem, which is formulated as an NP-hard combinatorial optimization with maximizing the success rate of matching as the objective.
Schleibaum and M{\"u}ller \cite{schleibaum2020human} advocate taking the determinants of user satisfaction and explainable matching decisions into consideration.
One of their future studies is to find out whether increasing the explainability can improve user satisfaction level or not.}

\revise{It is worth mentioning that many ML-based collective matching strategies take advantage of the Kuhn-Munkres (KM) bipartite matching algorithm as a component of their decision-making pipelines \cite{jonker1986improving}.
Drivers and riders are usually regarded as the two sets of vertices in the target bipartite graph.
To guide the matching between ride requests and ride offers,  Guo et al.~\cite{guo2020spatiotemporal} propose spatial-temporal Thermo, which is used to reflect the demand density of different places and times.
They use Random Forest Regression \cite{breiman2001random} to map multiple features of spatial, temporal, and meteorological dimensions to Thermo.
The weight of each pair of driver and rider in the bipartite graph is estimated by Thermo.
A KM algorithm is then used to calculate the final matching decisions according to the constructed bipartite graph.
Similarly, Xu et al.~\cite{xu2018large} derive their matching decisions using the KM algorithm.
In contrast to \cite{guo2020spatiotemporal}, Xu et al.~\cite{xu2018large} leverage a policy evaluation algorithm to learn a value function which maps each pair of driver and rider to a score.
The KM algorithm calculates the final matching between drivers and riders based on the scores.
Guo and Xu \cite{guo2020deep} also conduct the matching planning using the KM algorithm.
The weight between each pair of driver and rider is obtained from a value function learned by a convolutional neural network-based Double Q-learning (Double DQN) algorithm \cite{van2016deep}.}

\subsubsection{Distributed Matching}
\revise{RL is also a powerful technique for distributed matching \cite{sutton1999reinforcement}.
Gu{\'e}riau and Dusparic \cite{gueriau2018samod} use the Q-learning algorithm to train a policy for each agent (driver) to choose the pickup or rebalancing action based on the environment state, including the status of itself and current distribution of supply and demand.
If pickup action is chosen, then the vehicle will go and pick up the nearest rider.
In their follow-up work \cite{gueriau2020shared}, they extend the method to consider traffic congestion when agents are making decisions.
Wang et al.~\cite{wang2018deep} propose to use the DQN \cite{mnih2015human}, in which a deep neural network is employed to estimate the state-action value function from a single driver's perspective.
Many methods of distributed matching allow the decisions to be determined individually while the matching policy is trained collectively.
For example, De Lima et al.~\cite{de2020efficient} follow the QMIX framework proposed in \cite{rashid2018qmix}, in which the coordinated planning policies are trained by learning a joint action-value function for multiple vehicles and riders aiming at optimizing a global objective.
In the execution process, the matching decision of each vehicle is made in a distributed  manner following its own component in the learned action-value function.
By ensuring the monotonicity of the relationship between the global action-value and the action-value of each passenger, the objectives of distributed planning decisions are ensured to coincide with the centralized decisions during the training process.
Similar to \cite{de2020efficient}, Li et al.~\cite{li2019efficient} adopt the framework where the matching policy is trained in a centralized manner and executed in a distributed manner.
Specifically, they adopt the actor-critic RL framework, where actor and critic are two different networks used to decide and evaluate the action for each driver, respectively.
The coordination among drivers in the matching policy is enabled by the critic network.
It adopts the mean field approximation to model the interactions of drivers by calculating an average on the actions taken by their neighborhoods, which is then considered in the process of evaluating each driver's action.
\revise{In \cite{zhou2019multi}, another centralized training process is proposed, in which a Kullback–Leibler divergence optimization is used to balance the supply and demand and to enable coordination among the vehicles.}
In the execution phase, each driver chooses an action based on their own action-value functions.}

\revise{Some distributed matching strategies leverage other ML techniques.
They mostly determine the matching decisions based on the similarities between the riders and drivers in ride-hailing.
For example, Bicocchi and Mamei \cite{bicocchi2014investigating} use the bag-of-words model to summarize users' frequently visited places as vector representations, which are then fed to the Latent Dirichlet Allocation (LDA) \cite{blei2003latent} model to identify their patterns of daily travel routine behaviors.
Given a rider or a driver, his/her potential participants of shared-rides can be found by calculating the similarities between his/her daily travel routine and those of the other riders and drivers.
Lasmar et al.~\cite{lasmar2019rsrs} propose to leverage a multi-layer Perceptron model to learn user preferences based on their responses to the questionnaires.
For each rider, a ranking list of potential partners for shared-rides is generated according to the similarities between the predicted preferences of her/him and other riders.}

\subsection{Repositioning}
\label{sec:review-repositioning}
\subsubsection{Collective Repositioning}
\revise{Some collective repositioning methods leverage RL techniques.
Ride-hailing repositioning for electric vehicles is studied in \cite{liang2020mobility, tang2020online}, in which the state of charge of the electric vehicles is an important factor to be considered.
Liang et al.~\cite{liang2020mobility} develop a solution method utilizing deep RL combined with binary linear programming to obtain a regional joint planning policy for electric vehicles with their state of charge considered.
Using binary linear programming, each vehicle repositioning action is modeled as a binary decision variable, and its weight in the objective is obtained by the value function learned by the policy iteration method.
Similarly, Tang et al.~\cite{tang2020online} also combine RL with combinatorial optimization, in which the RL learned policy is used to advise decision making in the optimization step.
Liang et al.~\cite{liang2021integrated} adopt temporal-difference (TD) learning to obtain action-value function.
Different from \cite{de2020efficient}, the settings in \cite{liang2021integrated} do not allow factorization of the joint action-value function into individual ones while guaranteeing global maximization.
Thus, they formulate two linear programming instances to collectively find the decisions for the vehicles.
To improve the stability of the training process in RL, Fluri et al.~\cite{fluri2019learning} propose a cascading multi-level learning model.
In this model, the area concerned is split in halves as the number of levels of learning increases.
The policy training process proceeds in a top-down manner, i.e., from less to more fine-grained area partitioning. 
The motivation behind is that the policy trained from a coarse level can serve as guidance to the finer levels, which avoids the instability caused by directly training a policy with a large state size (w.r.t. the number of regions).
Fluri et al.~\cite{fluri2019learning} propose to leverage the Lloyd K-means algorithm \cite{lloyd1982least} to partition the area concerned into multiple smaller regions.
Deng et al.~\cite{deng2020multi} leverage the Proximal Policy Optimization algorithm (PPO) \cite{schulman2017proximal} to learn the joint repositioning policy for vehicles, in which the value- and policy-function are approximated by neural networks.
Shi et al.~\cite{shi2019optimal} use Deep Deterministic Policy Gradient (DDPG) \cite{silver2014deterministic} to learn the grid-based multiple vehicles repositioning policy with the objective of total profits maximization.
\revise{In \cite{shou2020reward}, a mean-filed multi-agent RL approach is leveraged to collectively relocate the vehicles in ride-hailing.}}

\revise{Some other collective repositioning solutions leverage various ML techniques to predict future information of a ride-hailing system, which plays an important role in guiding the platforms to make better repositioning decisions \cite{chen2022h}.
Riley et al.~\cite{riley2020real} leverage Vector autoregression to forecast the future demand from region to region.
The predicted demand and current system status are then fed into two mixed-integer programming instances to find the desired distribution of vehicles and the assignment of vehicles to regions, respectively.
Iglesias et al.~\cite{iglesias2018data} use a Long Short-Term Memory (LSTM) neural network to predict the future ride requests for each pair of origin and destination within a certain time period.
The predicted information is then used as input to their proposed mixed-integer linear programming instance, which is solved to find the optimal rebalancing actions.
Xu et al.~\cite{xu2018taxi} use two LSTM-based and Mixture Density Network (MDN)-based models to predict the distributions of origins and destinations of future requests, respectively.
With a prediction on the distributions, the repositioning decisions are then obtained by solving a mixed-integer programming problem with total idle driving distance minimization as the objective.
Cheng et al.~\cite{cheng18taxis} leverage a multilevel logistic regression model to predict the likelihood of ride requests occurring at different times and places.
The online repositioning planning decisions of drivers are obtained by leveraging a centralized multi-period stochastic optimization model with both the real-time and predicted demand considered.
Li et al.~\cite{li2020data} and Gao et al.~\cite{gao2020learning} formulate the repositioning task as a two-stage stochastic programming problem.
The source of the stochasticity is the underlying uncertainty of the future demands, the probability distribution of which is obtained by kernel density estimation and a deep learning model combining the LSTM and MDN in \cite{li2020data} and \cite{gao2020learning}, respectively.
Pouls et al.~\cite{pouls2020idle} propose a forecast-driven repositioning solution framework, the core of which is a mixed-integer programming problem with the demand predictions as inputs.
Moreover, it is solved by an off-the-shelf solver called Gurobi \cite{gurobi}.
Note that, in practice, not all planning decisions can be successfully executed by the drivers at the end.
Xu et al.~\cite{xu2020recommender} take the first step to predict the failure possibility of repositioning tasks in the decision-making process, including situations where drivers disobey the planning or end up being unmatched for an unexpectedly long time even though they follow the repositioning planning decisions accordingly.
In the latter case, drivers will be compensated.
The failure rate of each repositioning task is predicted by XGBoost \cite{chen2016xgboost} with both driver- and environment-related features as inputs.
\revise{The problem of multi-vehicle collaboration optimization aiming at maximizing the platform's profit is converted into a minimum cost flow problem, which is solved by an off-the-shelf method called GNU Linear Programming Kit (GLPK) \cite{makhorin2008glpk}. }}

\subsubsection{Distributed Repositioning}
Geographical regions or grids (i.e., abstracts of individual locations) are usually used to model the road networks in the problem of ride-hailing repositioning.
Different from most of the repositioning methods (e.g., \cite{lin2018efficient, riley2020real, ke2019optimizing, li2019efficient, zhou2019multi}) in which the region of interest is divided into predefined and static geographic zones, Castagna et al.~\cite{castagna2020demand, castagna2021demand} leverage the Expectation-Maximization clustering algorithm to derive zones for rebalancing vehicles in an online manner.
They leverage the Proximal Policy Optimization algorithm (PPO) \cite{schulman2017proximal} to train a policy for each vehicle to decide whether to make a pick-up, drop-off, or repositioning action.
Specifically, similar to \cite{tang2021value}, the repositioning destination is also sampled from a probability distribution over all potential positions, which is determined by the number of unserved requests.
Different from \cite{castagna2020demand, castagna2021demand}, Verma et al.~\cite{verma2017augmenting} propose an iterative method to dynamically split the zones based on their expected revenue (Q-values).
The iterative splitting process does not terminate until the historical data is exhausted for the Q-values learning.
\revise{Different from most of the works that model the drivers as agents, Jin et al.~\cite{jin2019coride} regard each geographical region as an agent.}
By hierarchically partitioning the target areas into regions with different granularities, they perform hierarchical RL where the multi-head attention mechanism is used to capture the impacts among the neighboring agents.
Guo et al.~\cite{guo2021multi} try various methods (e.g., Support Vector Regression, Random Forest Regression, and k-Nearest Neighbors regression) to predict future demand density, which is then used to evaluate each region for their spatial-temporal value.
Each available vehicle chooses to stay still or relocate to a neighbor region in a probabilistic manner based on their spatial-temporal values, which can help avoid over-saturation of supply.
In \cite{provoostdemandprop}, the region of interest is represented as a graph.
They build two neural networks to predict the demand on vertices and the passenger flows on edges, respectively.
The proposed repositioning algorithm aims at satisfying the demand on edges in the decreasing order with the nearest vehicles found by backward traversing.

\revise{However, in spite of the various grid-based methods as discussed in most of the related works mentioned above, e.g., \cite{lin2018efficient, guo2021multi}, Jiao et al.~\cite{jiao20deep, jiao2021real} argue that grid-based repositioning policies are not satisfactory in practice because of the excessively-simplified and overlooked non-stationarity in the environment caused by the dynamic environment and the large number of vehicles when coarse-grained region-wise decisions are considered.}
They put forward the process of carrying out repositioning 
in industrial production by combining offline learning, i.e., batch RL, and online planning stages, i.e., decision-time planning \cite{sutton1999reinforcement}.
To counter the issues of coarse-grained decisions, Kim and Kim \cite{kim2020optimizing} uses a graph to model the road networks which is more realistic.
They build a Graph Neural Network to predict the future demands.
The repositioning destination of each driver is decided greedily based on a function of the predicted demand, the number of excessive vehicles, and the distance information to each candidate position.

\revise{Some other works also spend special effort on tackling the non-stationarity.
With the observation that the actions of drivers are independent (based on self interests), Chaudhari et al.~\cite{chaudhari2020learn} propose a vanilla RL framework where each driver, based on a probabilistic value denoting the extent to which coordination is needed, stochastically chooses to perform an action guided by the independent or coordinated policy.
Note that, although vehicles execute repositioning decisions sequentially in this solution framework, coordination in the latter policy is explicitly considered by solving a minimum cost flow problem for the optimal rebalancing flow of vehicles among all the regions (which is similar to \cite{xu2020recommender}).
In addition, the independence between different repositioning policies learned by the drivers concurrently also contributes to the non-stationarity of the environment.
In this regard, Verma et al.~\cite{verma2019entropy} propose a method for each driver to learn the information of other vehicles in order to make a better planning decision.}
The principle of maximum entropy \cite{jaynes1957information} is leveraged to improve the predictability of the distribution of drivers even with only limited knowledge available, e.g., the local density of supply.
To tackle the non-stationary challenge in online ride-hailing as well as the catastrophic forgetting of RL \cite{kemker2018measuring}, Haliem et al.~\cite{haliem2020adapool, haliem2021adapool} propose to learn multiple repositioning policies to deal with different contexts of environments (e.g., peak/non-peak hours and weekends/weekdays).
When changes in the distribution of experiences are identified by their proposed change point detection algorithm, switching among those different policies is enabled so as to enhance adaptability to the dynamic environment.
Lei et al.~\cite{lei2019optimal} define the concept of stochastic relocation matrix.
The element in the $i$-th row and the $j$-th column within the matrix represents the probability that an empty vehicle located in the $i$-th region should relocate to the $j$-th region.
\revise{To circumvent the curse of dimensionality, they leverage low-rank approximation to project the original matrix onto a low-dimensional vector.}
They propose a deep convolution-LSTM model to learn how to predict the approximation vector based on the system status.
To alleviate the instability of the state-value function approximator caused by the large scale of its states, Tang et al.~\cite{tang2019deep} propose to bound its outputs by regularizing its worst-case variation w.r.t. 
changes in its inputs (i.e., states).
Transfer learning proposed in \cite{wang2018deep} is applied to increase the adaptability of the trained model across different cities.

\revise{Besides the traditional ML techniques discussed above, RL, being another well-known technique for decision making in non-stationary environments, has been a key technology in distributed repositioning \cite{khetarpal2022towards,xie2021deep,mao2021near}.}
\revise{Liu et al.~\cite{liu2022deep} propose a single-agent deep RL approach which relocates vacant vehicles to regions with a large demand gap in advance.}
Nguyen et al.~\cite{nguyen2018policy} propose to use the RL framework to train a homogeneous repositioning policy for all agents, i.e., vehicles.
\revise{The policy is trained in a centralized manner with collective behaviors of drivers considered while executing in a distributed manner.}
He and Shin \cite{he2019spatio} leverage Double DQN with their proposed spatial-temporal capsule-based neural network as the state-action value approximator.
The inputs of the network proposed include the location of the vehicle to be relocated, distribution of other vehicles and riders, ride preferences, and some external factors that have impacts on supply and demand, e.g., weather conditions and holiday events.
With all those information processed, the estimated value for each candidate position given the current state of the target vehicle is obtained, and the final decision can be decided in a probabilistic manner.
A more elaborate analysis is presented in their follow-up study \cite{he2020spatio}.
Yu et al.~\cite{yu2019markov} formulate the single-vehicle repositioning planning problem as a Markov Decision Process.
They propose to leverage parallelized matrix operations to re-formulate the Bellman equation \cite{sutton1999reinforcement}, thus reducing the computational complexity in finding optimal planning policy.
Multi-hop ride-hailing repositioning is considered in \cite{singh2019reinforcement, singh2021distributed}.
Similar to \cite{al2019deeppool}, they predict the number of vehicles in each region for certain time slots ahead of time using an estimated time of arrival (ETA) model.
Double DQN is adopted for each vehicle to choose the best neighbor region to move forward based on the current status of all the vehicles and the predicted demand and supply.
    
\subsection{Joint Matching and Repositioning}
\label{sec:review-joint}
In this part, we review methods that jointly optimize matching and repositioning with ML techniques. 
Note that all of them belong to the category of distributed planning. The research works in this part leverage RL to guide the decision making process.
Different from the review given by Qin et al.~\cite{qin2021reinforcement}, we focus on the works that jointly decide matching and repositioning.

Haliem et al.~\cite{haliem2020distributed-a, haliem2021distributed} propose to consider both the matching and repositioning in the ride-hailing planning process.
In their ride-hailing systems, each vehicle conduct initial matching by greedily searching the nearest requests, after which an insertion-based method is used to finalize the potential request list.
Then each driver, based on the value function learned by the DQN, weighs the requests in the final list.
The riders who receive those proposed ride offers can decide whether to accept the offers and join the trips where shared-rides are allowed.
The trips can be solo-ride or shared-ride.
Drivers are repositioned in parallel with the matching process.
Each driver takes actions indicated by his/her trained RL agent, i.e., the decision-making policy, independently.
Their proposed solution framework learns an optimal policy for each driver as opposed to those RL-based methods with collective planning scheme where a central policy is used, e.g., \cite{oda2018movi}.
Note that, in some works, although each driver makes decisions independently (e.g., \cite{haliem2020distributed-a, haliem2021distributed}), all drivers share one trained policy (e.g., \cite{manchella2020passgoodpool, manchella2021flexpool}).
Manchella et al.~\cite{manchella2020passgoodpool, manchella2021flexpool} propose to collectively optimize the system objectives, e.g., minimizing the waiting times and routing times.
Nevertheless, they allow distributed inference at the level of individual drivers. 
Their proposed model can be used by each vehicle independently.
It helps decrease computational costs associated with the growth of distributed systems. 
Specifically, they utilize a Double DQN with the experience relay mechanism.
Their model learns a probabilistic dependence between drivers' actions and the reward function.
The trained policy indicates a destination for each driver if s/he is not matched with any rider according to their proposed heuristic matching algorithm. 
Similar to \cite{xu2020highly, singh2019reinforcement, singh2021distributed}, multi-hop transit is enabled in their solutions.
Wang et al.~\cite{wang2018deep} model the matching and repositioning problems as a Markov Decision Process and propose learning solutions based on DQNs to optimize the trained policy for the drivers.
\revise{Their solution uses a temporal and spatial expanded action search strategy to accommodate the scenarios where there is only sparse training data, e.g., certain remote regions in the middle of the night.}
Besides, to increase the learning adaptability and
efficiency, they propose to use a transfer learning method to leverage the knowledge across both spatial and temporal spaces.

Besides \cite{haliem2020distributed-a, haliem2021distributed, manchella2020passgoodpool, manchella2021flexpool, wang2018deep},
DQN is used in other works as well, e.g., \cite{al2019deeppool, guo2022deep, tang2021value, li2020balancing}.
In \cite{al2019deeppool}, each vehicle decides its action by learning the impact of its action on the reward using a DQN model without coordinating with other vehicles.
In \cite{guo2022deep}, the vehicle repositioning procedure is formulated as a Markov Decision Process.
By sampling the future riders based on the historical probability distribution, the proactive relocation of vehicles is realized via a deep RL framework, which is composed of a Convolutional Neural Network and a Double DQN module. \revise{Then a request-vehicle assignment scheme is presented based on the value function attained from the vehicle repositioning process.}
\revise{Similarly, Tang et al.~\cite{tang2021value} propose a planning framework for tackling both the matching and repositioning tasks, the core of which is a unified value function which is trained offline using abundant historical data and is updated during the online phase.}
With the value function learned, the matching problem is then solved by the method proposed in \cite{xu2018large}, while the reposition destination of each idle vehicle is determined in a probabilistic manner following the distribution given by the discounted long-term values of all the candidate positions.
Li and Allan \cite{li2020balancing} also leverage a global value function for both the tasks of matching and repositioning, which is learned by the value iteration algorithm with historical data of ride requests.

	\begin{table*}[tp]
\centering
{\color{black}\begin{threeparttable}
\caption{\revise{Summary of open-source trip related data sets.}}
\begin{tabular}{@{}C{2.2cm}C{2.4cm}cccm{3.8cm}@{}}
\toprule
\textbf{Source}& \textbf{City} &\textbf{Type} & \textbf{\#~of Records} & \textbf{Time} & \multicolumn{1}{>{\centering\arraybackslash}m{3.8cm}}{\textbf{Data Features}}\\ \midrule
\midrule
Uber Pickups \cite{p668-gy46-22} & New York, United States& Pickups &20M & \begin{tabular}[c]{@{}c@{}}April -- September 2014,\\January -- June 2015 \end{tabular} & Timestamp and location.\\ \midrule
NYC TLC \cite{tlctrip} & New York, United States& Ride request &500M&January 2009 -- July 2021 &Passenger count, start time, end time, origin, destination, distance, fee, etc.\\ \midrule
DiDi GAIA \cite{gaiadata}& Haikou and Chengdu, China& Ride requests  &18M&\begin{tabular}[c]{@{}c@{}}November 2016,\\May -- October 2017\end{tabular}&Trip ID, fee, start time, end time, origin, destination, etc.\\ \midrule
Chicago Data Portal \cite{chicagoalll} & Chicago, United States & Ride requests &199M &January 2013 -- December 2021&Trip ID, taxi ID, start time, end time, fee, origin, destination, etc.\\ \midrule
T-drive \cite{tdrive, yuan2010t, yuan2011driving} & Beijing, China& Trajectories &15M&February 2008&Taxi ID, location, and timestamp.\\ \midrule
GeoLife \cite{geolifetraj, zheng2008learning, zheng2008understanding, zheng2010understanding} & Beijing, China& Trajectories &18,670&April 2007 -- August 2012 &Timestamp, location, transportation mode, etc. \\ \midrule
Beijing Taxi Trajectories \cite{lian2018one} & Beijing, China & Trajectories &129M&May 2019& Taxi ID, location, timestamp, speed, occupancy indicator, etc.\\ \midrule
DiDi GAIA \cite{gaiadata} & Chengdu and Xi'an, China &Trajectories &-\tnote{*}&-&\multicolumn{1}{>{\centering\arraybackslash}m{3.8cm}}{\textbf{-}}\\ \midrule
ECML PKDD 2015 \cite{portokaggle} & Porto, Portugal  & Trajectories  & 2M&July 2013 -- June 2014 &Trip ID, taxi ID, the sequence of locations of the trajectory, etc.\\ \midrule
CRAWDAD EPFL \cite{c7j010-22}  & San Francisco, United States & Trajectories  &11M&May -- June 2008& Taxi ID, location, timestamp, and occupancy indicator.\\\midrule
CRAWDAD Roma \cite{c7qc7m-22}  & Roma, Italy & Trajectories  &22M&February -- March 2014& Taxi ID, location, and timestamp.\\\midrule
Jeju Vehicular Trajectories \cite{y8vk-wj40-22}  & Jeju, South Korea & Trajectories  &8M& - & Vehicle ID, location, speed, lane, etc.\\\midrule
Grab-Posisi \cite{grabsource, huang2019grab}  &Singapore and Jakarta, Indonesia& Trajectories  &84,000&April 2019&Trajectory ID, location, timestamp, speed, etc.\\\midrule
Shanghai Taxi Trajectories \cite{2877-mk46-19, liu2020optimization}  &Shanghai, China& Trajectories  &61M&April 1, 2018&Taxi ID, location, timestamp, speed, occupancy indicator, driving status, etc.\\\midrule
Foursqure \cite{foursquare} & Global & Check-ins &33M&April 2012 -- September 2013 & Venue ID, timestamp, location, etc.\\ \midrule
Brightkite \cite{brightkite} & - & Check-ins &4M&April 2008 -- October 2010&User ID, timestamp, location, etc.\\ \midrule
Gowalla \cite{gowalla} & -  & Check-ins &6M&February 2009 -- October 2010&User ID, timestamp, location, etc.\\ \midrule
LTA of Singapore \cite{singaporedata} & Singapore & \begin{tabular}[c]{@{}c@{}}Real-time\\locations\end{tabular}  &-&-&Location. \\\midrule
Uber \cite{traveltimedata} &  Global &Travel times  &-&-&Origin, destination, travel time, etc.\\\bottomrule
\end{tabular}
\label{table:data}
\begin{tablenotes}
\item[*] \revise{``-'' indicates that the corresponding information is not attainable.}
\end{tablenotes}
\end{threeparttable}}
\end{table*}

\begin{table*}[t]
\centering
{\color{black}\begin{threeparttable}
\caption{Summary of the major simulators.}
\begin{tabular}{@{}cccm{9.6cm}@{}}
\toprule
\textbf{Name} & \textbf{Open-Source} & \begin{tabular}[c]{@{}c@{}}\textbf{Sample Applications}\\\textbf{in Ride-hailing} \end{tabular}& \multicolumn{1}{>{\centering\arraybackslash}m{9.6cm}}{\textbf{Description}} \\ \midrule
\midrule
 DiDi \cite{didisimulation} & \xmark &\cite{tang2021value}&An online simulation platform to evaluate matching and repositioning algorithms.  Evaluation is run with DiDi's real-world data.\\ \midrule
 AMoDeus \cite{ruch2018amodeus} & \cmark&\cite{ruch2020quantifying}&A tool that uses an agent-based transportation simulation framework to simulate arbitrarily configured mobility-on-demand systems with static/dynamic demand. It includes standard benchmark algorithms and a graphical user interface.\\ \midrule
 AMoD2 \cite{amod2} & \cmark&\cite{li2021optimal}&A high-capacity ride-sharing simulator. It uses map data and taxi data from Manhattan. Three matching algorithms and one simple rebalancing algorithm are implemented. \\ \midrule
 Mod-abm \cite{abm-1.0, abm-2.0} & \cmark&\cite{wen2017rebalancing}&A platform for simulation of large-scale mobility-on-demand operations. It supports city-level systems in any urban setting.\\ \midrule
 MATSim \cite{horni2016multi} & \cmark&\cite{tsao2019model}&An open-source framework for implementing large-scale agent-based transport simulations. It consists of several modules which can be combined or used in stand-alone mode, for demand-modeling, traffic flow simulation, re-planning; and a controller to iteratively run simulations, and methods to analyze outputs generated by the modules.\\ \midrule
 SUMO \cite{lopez2018microscopic} & \cmark&\cite{castagna2021multi,zhu2021shared}&A microscopic and space-continuous traffic simulation platform that is suitable for the generation, evaluation, and validation of traffic scenarios of real-world size. It supports road network customization and demand modeling.\\ \midrule
 CityFlow \cite{zhang2019cityflow} & \cmark&-\tnote{*}&A multi-agent RL environment for large scale city traffic scenario. It supports flexible definitions for road network and traffic flow. It provides faster simulation than SUMO.\\ \midrule
 STaRS \cite{ota2016stars} & \xmark &-&A simulation framework for analyzing diverse ride-sharing scenarios, considering the platforms' needs and constraints. Its real-time trip assignments utilize a linear optimization algorithm, efficient indexing, and parallelization for scalability.\\ \midrule
 STaRS+ \cite{mounesan2021fleet} & \xmark &-&A simulation framework based on an integer linear programming model, using heuristic optimization and a novel shortest-path caching scheme for scalability. It supports the simulation of full-city scale ride-sharing with meeting points.\\ \midrule
 UberSim \cite{khalil2022realistic} & \cmark & \cite{salman2023quantifying} &A digital twin transportation simulation model for Birmingham, Alabama, incorporating various transportation modes to analyze the impact of ride-hailing services on urban traffic. It supports policy learning through reinforcement learning.\\ \midrule
 NYC-Yellow-Taxi-V0 \cite{chaudhari2020learn} & \cmark&-&A multi-agent RL environment based on the OpenAI Gym environment \cite{openai-gym}, which offers a toolkit for developing and comparing RL-based ride-hailing fleet management algorithms.\\ \midrule
 SMART-eFlo \cite{liu2022smart} & \cmark & - &An integrated framework that combines the SUMO simulator with multi-agent Gym for reinforcement learning studies, enabling researchers to easily design traffic scenarios and implement RL algorithms for electric fleet management problems\\ \midrule
\end{tabular}
\label{table:simulator}
\begin{tablenotes}
\item[*] ``-'' indicates no sample application in ride-hailing using the corresponding simulator.
\end{tablenotes}
\end{threeparttable}}
\end{table*}

\section{Resources for empirical studies}
\label{sec:resource}
Real-world data are essential in studying ML-based ride-hailing planning, e.g., to train a proposed model.
Further, in order to deploy those proposed planning strategies in practice, it is necessary to utilize a ride-hailing service (process) simulator to validate their performance with real-world data.
\revise{In this section, we present publicly available 
open-source real-world data sets and several related simulators in Sec.~\ref{sec:resource-data} and Sec.~\ref{sec:resource-simulator}, respectively.}

\subsection{Data}
\label{sec:resource-data}
Historical records of ride requests are the most frequently used data in the literature.
New York City Taxi and Limousine Commission \cite{tlctrip} provides more than 11 years of records of ride requests with many data fields, e.g., the pickup and drop-off timestamps, pickup and drop-off locations, trip fare, and ride distance.
About 19 million Uber's pick-up records obtained from this data set 
are summarised in \cite{p668-gy46-22}.
Chicago Data Portal \cite{chicagoalll} provides a large data set consisting of trip records with various data fields recorded from 2013.
Ride request data of Haikou and Chengdu in China are publicly available in the DiDi GAIA program \cite{gaiadata}.
Besides, there are some other data sets that consist of people's check-in records, e.g., those collected from Foursquare \cite{foursquare}, Brightkite \cite{brightkite}, and Gowalla \cite{gowalla}.
They are informative in revealing ride demands as the data have recorded riders' target places they had traveled to.
In addition to the records of ride requests, some data of the supply side are also available.
\revise{Trajectory data recorded by sampling drivers' locations at a certain frequency are also commonly used in the literature, including trajectories recorded in Beijing (in three different data sets: T-drive \cite{tdrive, yuan2010t, yuan2011driving}, GeoLife \cite{geolifetraj, zheng2008learning, zheng2008understanding, zheng2010understanding}, and Beijing Taxi Trajectories \cite{lian2018one}), Chengdu \cite{gaiadata}, Xi'an \cite{gaiadata}, Porto \cite{portokaggle}, Jakarta \cite{grabsource, huang2019grab}, Singapore \cite{grabsource, huang2019grab}, San Francisco \cite{c7j010-22}, Roma \cite{c7qc7m-22}, Jeju \cite{y8vk-wj40-22}, and Shanghai \cite{2877-mk46-19, liu2020optimization}.}
The Land Transport Authority (LTA) of Singapore \cite{singaporedata} publicizes many APIs for accessing various kinds of transport-related data, including monthly statistics of taxi supply and real-time coordinates of all taxis that are currently available for hire. 
\revise{Besides, the travel times among different locations in various cities across the world can be obtained in \cite{traveltimedata}.}
The data sets mentioned above are summarized in Table.~\ref{table:data}.
\revise{Note that, although abundant public data sets are available for use, none of them record those ride requests that ended up unserved (because of, for example, excessive waiting time).}
If both the served and unserved ride requests are treated in the ride-hailing simulation, the simulation results would be more realistic than if only
the served ride requests are considered. 

In addition to the trip-related data sets mentioned above, there are several important public data sources that could provide good values
to ride-hailing planning studies.
\revise{OpenStreetMap \cite{openstreetmap} captures worldwide road networks, which can be easily accessed by APIs (e.g., OSMnx) using Python \cite{boeing2017osmnx}.}
Travel times and speeds information of road networks of several well known cities have been made available in \cite{traveltimedata}.
Finally, historical data of weather conditions can be found in \cite{weatherdata}.
Note that weather conditions can be considered in ride-hailing planning
as they could affect the traffic in a major way
\cite{he2019spatio}.

\subsection{Simulators}
\label{sec:resource-simulator}
With the aforementioned real-world data, proposed planning strategies can be evaluated through simulators.
DiDi \cite{didisimulation} makes public an industrial-level simulation platform recently, which is used in the KDD CUP 2020 \cite{kddcup20}. 
The platform evaluates submitted matching and repositioning strategies using real-world data from the GAIA program.
Wen \cite{abm-1.0} develops an agent-based modeling platform for simulating autonomous mobility-on-demand systems, which is later upgraded to a version with better scalability and extensibility \cite{abm-2.0}.
Based on \cite{abm-2.0}, a high-capacity on-demand ride-sharing simulator is proposed in \cite{amod2} with several built-in matching and repositioning algorithms.
Ruch et al.~\cite{ruch2018amodeus} have released an open-source simulator named AMoDeus for accurate and quantitative analysis of matching and repositioning algorithms in the ride-hailing system.
Examples of its usages can be found in \cite{fluri2019learning, carron2019scalable}.
AMoDeus is built upon the open-source microscopic multi-agent transportation simulation environment MATSim \cite{horni2016multi}.
MATSim is also used to evaluate ride-hailing planning strategies, a case study of which can be found in \cite{bischoff2016simulation}.
Besides MATSim, there are similar public traffic simulation tools, including SUMO \cite{lopez2018microscopic} and CityFlow \cite{zhang2019cityflow}.
A more detailed comparison and analysis of the performance of traffic simulation tools can be found in \cite{allan2015benchmark}.
\revise{STaRS is a scalable simulation framework that takes the needs and constraints of platforms into consideration \cite{ota2016stars}. 
STaRS+ extends STaRS to support the simulation of ride-sharing with meeting points \cite{mounesan2021fleet}.}
\revise{For those RL-based ride-hailing planning methods, their policies can be trained using CityFlow \cite{zhang2019cityflow}, UberSim \cite{khalil2022realistic}, NYC-Yellow-Taxi-V0 \cite{chaudhari2020learn}, or SMART-eFlo \cite{liu2022smart}.}
The aforementioned simulators are summarized in Table.~\ref{table:simulator}.

	\section{Future directions and challenges}
\label{sec:future}
\begin{figure*}[h]
	\centering
	\includegraphics[width=0.9\linewidth]{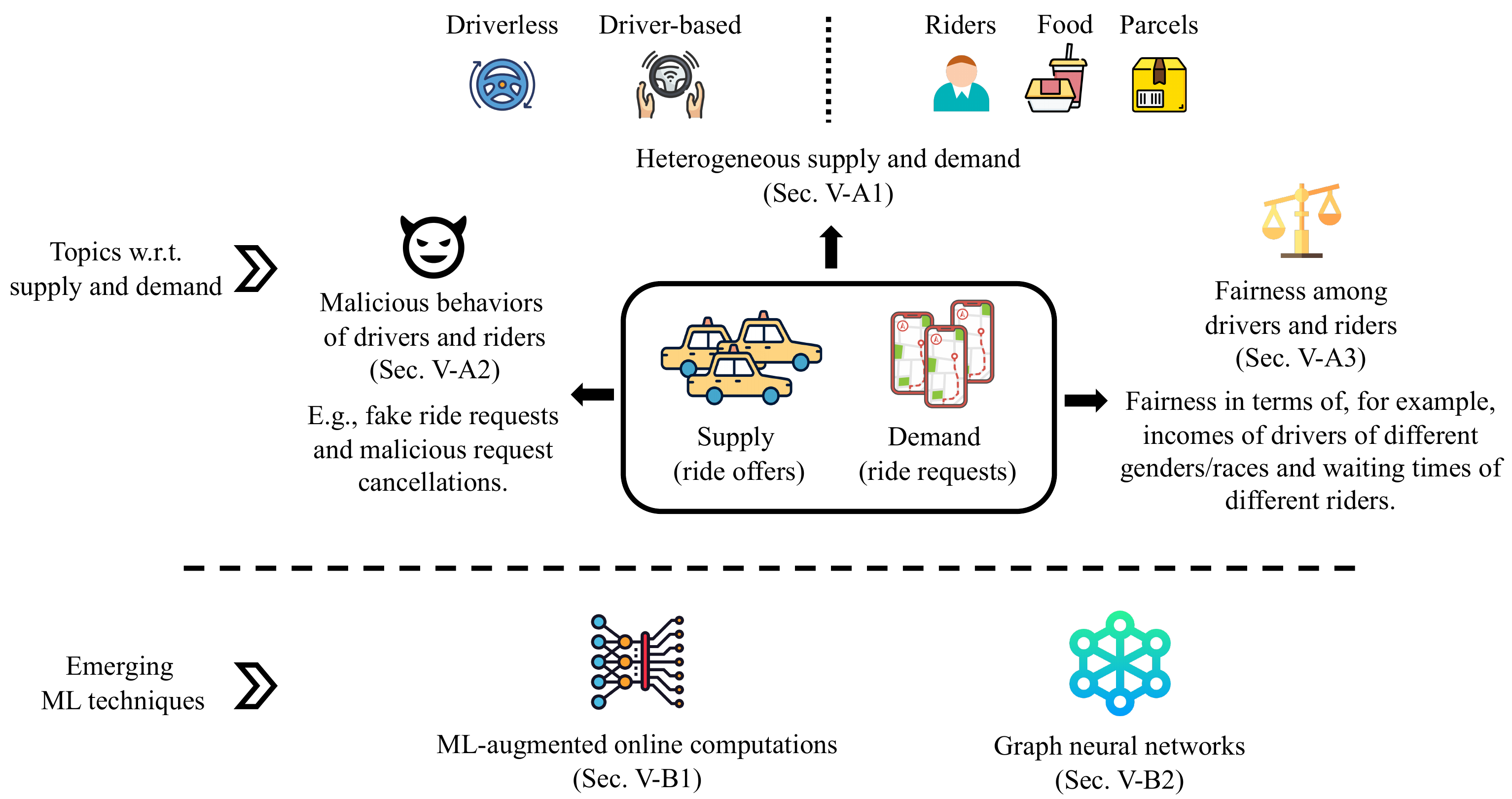}
    \caption{\revise{The future directions of ride-hailing planning.}}
	\label{fig:future}
\end{figure*}

\revise{In this section, we propose several future directions from the perspectives of ride-hailing supply (i.e., ride offers) and demand (i.e., ride requests) and emerging ML techniques in Secs.~\ref{sec:future-supplyanddemand} and \ref{sec:future-ml}, respectively.
\revise{The topics we present below appear well-motivated but are largely overlooked in existing ML-based ride-hailing planning literature.}
Fig.~\ref{fig:future} provides an outline of the proposed future directions.}

\subsection{\revise{Topics Regarding Supply and Demand}}
\label{sec:future-supplyanddemand}
\revise{In this part, we first propose to study the problem of ride-hailing planning with heterogeneous supply and demand.
Then, we discuss malicious behaviors and unfairness issues in ride-hailing planning.}

\subsubsection{\revise{Heterogeneous Supply and Demand}}
\emph{Heterogenous} supply and demand exist in ride-hailing planning. 
\revise{With emerging self-driving technologies, leveraging driverless vehicles in ride-hailing planning is a promising direction \cite{mao2020dispatch, mo2022modeling, fan2022joint}.}
In this regard, vehicles can be categorized into driverless or driver-based.
\revise{Driverless vehicles are more flexible in planning than driver-based ones since there is no restriction on working hours and regional concerns, and no issue with drivers’ preferences 
\cite{zhang2022autonomous}.} 
\revise{Heterogeneity is even more conspicuous in ride-hailing platforms such as DiDi, Uber, and Lyft as they provide ride services to not only the passengers but also food and parcel deliveries.}
\revise{Different types of deliverables have different requirements regarding 
the tolerable waiting time and detour distance.}
For instance, in contrast to the ordered takeout food, users tend to be more tolerant about the delivery time of receipt of parcels.

\revise{A possible line of research is ride-hailing planning with heterogeneous supply and demand.}
An example is to take advantage of passengers' availability in ride-hailing planning to achieve better performance as compared to the traditional \emph{homogeneous} one.
\revise{For instance,
consider the situation where a passenger is willing to act as the delivery agent for parcels (or even food items) that are heading the same direction or share the same destination as the passenger. As an incentive the passenger is properly compensated for. It might sound somewhat far-fetched, but passengers acting as courier agent has for decades been a common practice in air travel.}
\revise{Such heterogeneity in supply and demand, despite its potential benefits as can be imagined,
introduces additional challenges in 
ride-hailing planning.}
Specifically, more constraints on riders are added to the problem.
\revise{In the example above, we need to ensure that the rider would pick up and drop off the food items or parcels at prescribed locations quickly enough to satisfy all the deadlines or time tolerance.}
\revise{Obviously, the decision making process w.r.t.~matching would become more complex than the case of homogeneous ride-hailing.}
\revise{Repositioning is also more complicated since it is necessary to consider multiple distributions of various types of vehicles across different regions when trying to balance the supply and demand.}

\subsubsection{\revise{Malicious Behaviors of Drivers and Riders}}
\revise{Malicious behaviors of drivers and riders have been observed in ride-hailing systems.}
\revise{To incentivize more drivers to enrol in ride-hailing systems, platforms offer monetary rewards 
to drivers when they manage to obtain certain achievements, e.g.,
number of riders served reaching a respectable level.
To secure such an award, rogue drivers may abuse the mechanism by generating fake passenger accounts, ride requests, etc.}
\revise{Existing planning algorithms that are oblivious of such behaviors
would match the fake ride requests like normal ones, and thus reduce the number of vehicles that are available to the real requests.}
\revise{Malicious riders can also cancel their ride requests deliberately while the assigned drivers are on the way to their origins.}
Fink \cite{cnn2014uber} reveals that employees from one of the leading ride-hailing companies made more than 5,000 ride requests on their rival platform but canceled them at last in one month.
The planning algorithms can be deceived by these riders and allocate vehicle resources to meet their phony ride requests.
\revise{Such malicious behaviors impact the availability of drivers, and reduce the utility of drivers, platforms, and other honest riders.
They can lead to poor performance of existing methods as reviewed in Sec.~\ref{sec:review} in terms of total number of served requests and riders' waiting time, etc.}
\revise{To study the problem of ride-hailing planning in the presence of malicious behaviours of drivers and riders could be an important future direction.}

It is challenging to design such robust planning algorithms.
\revise{With no sign of misbehaviour,
malicious entities interact with ride-hailing platforms and make requests for rides. 
They may then subdue the planning strategies, e.g., the RL-based matching and repositioning policies through attacks such as blackbox adversarial attacks \cite{zhao2020blackbox}.
Therefore, possible malicious behaviors must be adequately dealt with in an ML-based planning strategy.
It complicates the design space of effective ride-hailing planning decisions.}

\subsubsection{\revise{Fairness Among Drivers and Riders}}
\revise{Fairness is about the equality of treatment.
There are evidences indicating that profit-oriented policies of ride-hailing platforms can lead to discrepancies in the incomes of drivers of different genders and/or different races \cite{cook2018gender, tjaden2018ride, wang2020disruptive}.}
As for the riders, platforms' planning decisions do not always coincide with the desirability of every rider \cite{wolfson2017fairness}. 
\revise{That is, some riders might be offered purposely a better treatment than the others, e.g., less waiting time or less detour distance.}
Efforts have been made in using non ML-based ride-hailing planning strategies to counter unfairness for drivers \cite{lesmana2019balancing, xu2020trade, nanda2020balancing} or riders \cite{wolfson2017fairness, suhr2019two, raman2021data}.
However, existing ML-based ride-hailing planning methods (as reviewed in Sec.~\ref{sec:review}) have largely overlooked the fairness issues.
ML is effective and allows us to make decisions intelligently by training models using real-world data. 
Unfairness therefore inevitably arises as the real-world data have recorded and encoded unfairness happening in the real world. 
\revise{Such blemishes in the data are also learnt in the model training process \cite{barocas2017fairness}, thus causing the inequality of treatments in subsequent ride-hailing planning.}

\revise{The main challenge of designing ML-based ride-hailing planning strategies while ensuring fairness on the part of the drivers and riders is that the system utility, e.g., the profit of platforms, could run into conflict with the fairness metrics \cite{xu2020trade}, e.g., the per unit time profit of different drivers.}
Specifically, when the system utility is improved, fairness may worsen, and vice versa.
\revise{It is important and challenging to find a strategy with a trade-off between the utility and fairness that is as \emph{tight} as possible, meaning that the achieved utility (resp.~fairness) cannot be improved without making the fairness (resp.~utility) worse, i.e., the state of Pareto optimum.}
\revise{Further, when fairness among both riders and drivers is considered, many constraints need to be addressed, e.g., equal waiting times among riders and equal profits among different drivers.}
\revise{Clearly,
achieving a satisfactory system utility that upholds fairness is more challenging than the traditional ride-hailing problem.}

\subsection{\revise{Topics Regarding Emerging ML Techniques}}
\label{sec:future-ml}
\revise{Next, we shed light on some emerging ML techniques that can potentially improve the performance of ride-hailing planning, including ML augmented online computations and graph neural networks.}

\subsubsection{\revise{ML-Augmented Online Computation}}

\revise{As mentioned in Sec.~\ref{sec:background}, as ensuing supplies and demands are uncertain in general, the ride-hailing planning decisions need to be made in an online manner.
Online algorithms is a well-known paradigm for tackling this kind of online decision-making problems with performance guarantees.}
Some existing works have proposed online ride-hailing planning schemes, e.g., \cite{dickerson2018allocation, zhao2019preference, xu2020trade, nanda2020balancing, xu2020unified, lowalekar2020competitive}.
These works assume no knowledge about the future during the online process.
However, in the real world, some parts of such knowledge, e.g., the future distribution of ride requests, can be estimated via advanced ML techniques.
Thus, the assumption of zero knowledge made in the above works is too pessimistic in practice \cite{lykouris2018competitive}.
\revise{A fast-growing trend in addressing this issue is to augment online algorithms with ML predictions in order to achieve better performance, even in practical systems, while guaranteeing the worst-case performance 
\cite{lykouris2018competitive, purohit2018improving, anand2020customizing, NEURIPS2020_e834cb11, rohatgi2020near, mitzenmacher2020algorithms}. }

The challenges of applying such techniques are mainly twofold.
First, we need to decide what predictions to rely on.
Different predictions lead to different online decision-making processes and outcomes.
Existing ML-augmented online algorithms use problem-specific predictions. 
\revise{For example, online ski-rental uses the predicted weather conditions to assist the decision making process which decides whether to rent or buy the ski so as to minimize the cost \cite{purohit2018improving}.}
\revise{Such problem settings and the corresponding algorithmic analysis are much more than a simple extension of the original ride-hailing planning problem.}
Second, predictions can easily be inaccurate, which complicates the design for a planning scheme that requires performance guarantee.
A desirable online planning algorithm should be one that is unaware of and make no assumptions on the prediction accuracy.
Specifically, the performance of the algorithm is expected to be close to the best offline oracle for our problem when the predictions are accurate.
When the predictions are completely inaccurate, its performance should be close to that of the best online planning algorithm without predictions.

\subsubsection{\revise{Graph Neural Networks}}
Existing works of ML-based ride-hailing planning usually use coarse-grained abstractions of geographical locations, i.e., grid-based formulation, to model road networks \cite{ke2019optimizing, li2019efficient, lin2018efficient, zhou2019multi}.
However, grid-level planning is not practical \cite{verma2017augmenting}.
\revise{For example, when a driver is assigned to a certain grid according to the decision made by an algorithm, the driver has no idea as to which specific location inside the grid s/he should head towards.}
This is particularly problematic when a grid corresponds to a large region in the real world.
Instead, we can use graph models for ride-hailing planning  \cite{guo2020spatiotemporal, kim2020optimizing}.
\revise{Road networks can be modeled as graphs.
In a road graph, nodes are geographical locations, and edges are travel paths.}
Additional information, such as the number of riders and drivers, and the travel costs on paths, can be used as representations of the nodes and edges.
Such graphs are represented in a non-Euclidean space which is difficult for knowledge extraction using traditional ML techniques.
In response, graph neural networks (GNNs), the deep learning models that are capable of tackling graph-related tasks, were proposed \cite{wu2020comprehensive}.
We propose to use GNN to solve the ride-hailing planning problem in an end-to-end manner.  
\revise{Specifically, we can model road networks via graphs, design a GNN-based model to extract features from the graphs, and then map them to planning decisions.}
\revise{There are existing works that have employed GNNs in the transportation area, e.g., to calculate traffic predictions \cite{li2018diffusion, yu2018spatio} and demand predictions \cite{yao2018deep, kim2020idle}.
\revise{Only a limited number of works however have used GNN in ride-hailing planning, e.g., \cite{kim2020optimizing, li2022decentralized}.}
\revise{The algorithm in \cite{kim2020optimizing} can only match drivers and riders that are located on the same road segment, which restricts many other drivers and riders from being matched, and thus could lead to poor performance of their proposed method in terms of measures such as the total waiting times of riders.
As for the approach proposed in \cite{li2022decentralized}, it models the drivers and riders as nodes in a graph and but overlooks the road conditions, e.g., critical road intersections, popular pick-up locations, and drop-off locations, which may lead to planning decisions that are too coarse.}
}
\revise{To achieve better performance, we need to design new GNN-based models for ride-hailing planning that can address the issue more effectively.}

It is challenging to use GNNs in ride-hailing planning.
\revise{First, we need to identify how to model real-world information, e.g., the pickup and drop-off points, the number of riders and drivers, and the travel costs, as nodes and edges and to encode in them sufficient knowledge that can be learned by the GNN-based model to guide the planning process accordingly.}
The core of a GNN-based model is the \emph{information diffusion scheme} which aggregates and extracts knowledge from nodes and edges.
In ride-hailing, the temporal and spatial relations among different nodes and edges, e.g., the travel costs among various locations in different time periods, are critical to the planning decisions.
\revise{To improve the capability of a GNN-based model in learning such relations, an elaborately designed information diffusion scheme is desired.}

	\section{Conclusion}
\label{sec:conclusion}
Ride hailing is an important and indispensable aspect of living in developed countries and modern metropoles. In this article, we give a comprehensive overview of the literature on using ML to improve and enhance ride-hailing matching and repositioning.
\revise{We propose a 
taxonomy based on which we divide the current strategies into different categories according to the types of the planning tasks and the nature of the methods.}
Based on the proposed taxonomy, we compare and summarize the most essential literature relevant to the topic.
Resources of data sets and simulators are reviewed and compared, which should be most useful for researchers conducting empirical studies.
At last, we discuss several future directions and their challenges.
	
\bibliographystyle{IEEEtran}

\begin{thebibliography}{100}
\providecommand{\url}[1]{#1}
\csname url@samestyle\endcsname
\providecommand{\newblock}{\relax}
\providecommand{\bibinfo}[2]{#2}
\providecommand{\BIBentrySTDinterwordspacing}{\spaceskip=0pt\relax}
\providecommand{\BIBentryALTinterwordstretchfactor}{4}
\providecommand{\BIBentryALTinterwordspacing}{\spaceskip=\fontdimen2\font plus
\BIBentryALTinterwordstretchfactor\fontdimen3\font minus
  \fontdimen4\font\relax}
\providecommand{\BIBforeignlanguage}[2]{{%
\expandafter\ifx\csname l@#1\endcsname\relax
\typeout{** WARNING: IEEEtran.bst: No hyphenation pattern has been}%
\typeout{** loaded for the language `#1'. Using the pattern for}%
\typeout{** the default language instead.}%
\else
\language=\csname l@#1\endcsname
\fi
#2}}
\providecommand{\BIBdecl}{\relax}
\BIBdecl

\bibitem{jye19transport}
J.~Ye, ``Transportation: A data driven approach,'' in \emph{Proc.~of ACM
  SIGKDD}, 2019.

\bibitem{iqbal22uber}
``Uber revenue and usage statistics (2022),''
  \url{https://www.businessofapps.com/data/uber-statistics/}, visted on May 25,
  2022.

\bibitem{ma2017morning}
R.~Ma and H.~Zhang, ``The morning commute problem with ridesharing and dynamic
  parking charges,'' \emph{Transportation Research Part B: Methodological},
  vol. 106, pp. 345--374, 2017.

\bibitem{cramer2016disruptive}
J.~Cramer and A.~B. Krueger, ``Disruptive change in the taxi business: The case
  of uber,'' \emph{American Economic Review}, vol. 106, no.~5, pp. 177--82,
  2016.

\bibitem{feng2020we}
G.~Feng, G.~Kong, and Z.~Wang, ``We are on the way: Analysis of on-demand
  ride-hailing systems,'' \emph{Manufacturing \& Service Operations
  Management}, 2020.

\bibitem{hyland2020operational}
M.~Hyland and H.~S. Mahmassani, ``Operational benefits and challenges of
  shared-ride automated mobility-on-demand services,'' \emph{Transportation
  Research Part A: Policy and Practice}, vol. 134, pp. 251--270, 2020.

\bibitem{kpmg20smart}
``Smart cities and its relevance to mobility,''
  \url{https://assets.kpmg/content/dam/kpmg/za/pdf/pdf2020/smart-cities-and-its-relevance-to-mobility.pdf},
  visted on May 25, 2022.

\bibitem{taiebat2022sharing}
M.~Taiebat, E.~Amini, and M.~Xu, ``Sharing behavior in ride-hailing trips: A
  machine learning inference approach,'' \emph{Transportation Research Part D:
  Transport and Environment}, vol. 103, p. 103166, 2022.

\bibitem{holler2018deep}
J.~Holler, Z.~Qin, X.~Tang, Y.~Jiao, T.~Jin, S.~Singh, C.~Wang, and J.~Ye,
  ``Deep q-learning approaches to dynamic multi-driver dispatching and
  repositioning,'' in \emph{Proc.~of NeurIPS Deep RL Workshop}, 2018.

\bibitem{holler2019deep}
J.~Holler, R.~Vuorio, Z.~Qin, X.~Tang, Y.~Jiao, T.~Jin, S.~Singh, C.~Wang, and
  J.~Ye, ``Deep reinforcement learning for multi-driver vehicle dispatching and
  repositioning problem,'' in \emph{Proc.~of IEEE ICDM}, 2019.

\bibitem{qin2019deep}
Z.~T. Qin, X.~Tang, Y.~Jiao, F.~Zhang, C.~Wang, and Q.~T. Li, ``Deep
  reinforcement learning for ride-sharing dispatching and repositioning,'' in
  \emph{Proc.~of IJCAI}, 2019.

\bibitem{jin2019coride}
J.~Jin, M.~Zhou, W.~Zhang, M.~Li, Z.~Guo, Z.~Qin, Y.~Jiao, X.~Tang, C.~Wang,
  J.~Wang \emph{et~al.}, ``Coride: joint order dispatching and fleet management
  for multi-scale ride-hailing platforms,'' in \emph{Proc.~of ACM CIKM}, 2019.

\bibitem{tafreshian2020frontiers}
A.~Tafreshian, N.~Masoud, and Y.~Yin, ``Frontiers in service science: Ride
  matching for peer-to-peer ride sharing: A review and future directions,''
  \emph{Service Science}, vol.~12, no. 2-3, pp. 44--60, 2020.

\bibitem{agatz2011dynamic}
N.~Agatz, A.~L. Erera, M.~W. Savelsbergh, and X.~Wang, ``Dynamic ride-sharing:
  A simulation study in metro atlanta,'' \emph{Procedia-Social and Behavioral
  Sciences}, vol.~17, pp. 532--550, 2011.

\bibitem{lloret2017peer}
R.~Lloret-Batlle, N.~Masoud, and D.~Nam, ``Peer-to-peer ridesharing with
  ride-back on high-occupancy-vehicle lanes: Toward a practical alternative
  mode for daily commuting,'' \emph{Transportation Research Record}, vol. 2668,
  no.~1, pp. 21--28, 2017.

\bibitem{ta2017efficient}
N.~Ta, G.~Li, T.~Zhao, J.~Feng, H.~Ma, and Z.~Gong, ``An efficient ride-sharing
  framework for maximizing shared route,'' \emph{IEEE Transactions on Knowledge
  and Data Engineering}, vol.~30, no.~2, pp. 219--233, 2017.

\bibitem{long2018ride}
J.~Long, W.~Tan, W.~Szeto, and Y.~Li, ``Ride-sharing with travel time
  uncertainty,'' \emph{Transportation Research Part B: Methodological}, vol.
  118, pp. 143--171, 2018.

\bibitem{stiglic2016making}
M.~Stiglic, N.~Agatz, M.~Savelsbergh, and M.~Gradisar, ``Making dynamic
  ride-sharing work: The impact of driver and rider flexibility,''
  \emph{Transportation Research Part E: Logistics and Transportation Review},
  vol.~91, pp. 190--207, 2016.

\bibitem{regue2016car2work}
R.~Regue, N.~Masoud, and W.~Recker, ``Car2work: Shared mobility concept to
  connect commuters with workplaces,'' \emph{Transportation Research Record},
  vol. 2542, no.~1, pp. 102--110, 2016.

\bibitem{bei2018algorithms}
X.~Bei and S.~Zhang, ``Algorithms for trip-vehicle assignment in
  ride-sharing,'' in \emph{Proc.~of AAAI}, 2018.

\bibitem{tamannaei2019carpooling}
M.~Tamannaei and I.~Irandoost, ``Carpooling problem: A new mathematical model,
  branch-and-bound, and heuristic beam search algorithm,'' \emph{Journal of
  Intelligent Transportation Systems}, vol.~23, no.~3, pp. 203--215, 2019.

\bibitem{noruzoliaee2022one}
M.~Noruzoliaee and B.~Zou, ``One-to-many matching and section-based formulation
  of autonomous ridesharing equilibrium,'' \emph{Transportation Research Part
  B: Methodological}, vol. 155, pp. 72--100, 2022.

\bibitem{masoud2017real}
N.~Masoud and R.~Jayakrishnan, ``A real-time algorithm to solve the
  peer-to-peer ride-matching problem in a flexible ridesharing system,''
  \emph{Transportation Research Part B: Methodological}, vol. 106, pp.
  218--236, 2017.

\bibitem{agatz2010sustainable}
N.~Agatz, A.~Erera, M.~Savelsbergh, and X.~Wang, ``Sustainable passenger
  transportation: Dynamic ride-sharing,'' \emph{ERIM Report Series Research in
  Management}, no. ERS-2010-010-LIS, 2010.

\bibitem{masoud2017decomposition}
N.~Masoud and R.~Jayakrishnan, ``A decomposition algorithm to solve the
  multi-hop peer-to-peer ride-matching problem,'' \emph{Transportation Research
  Part B: Methodological}, vol.~99, pp. 1--29, 2017.

\bibitem{alonso2017demand}
J.~Alonso-Mora, S.~Samaranayake, A.~Wallar, E.~Frazzoli, and D.~Rus,
  ``On-demand high-capacity ride-sharing via dynamic trip-vehicle assignment,''
  \emph{Proceedings of the National Academy of Sciences}, vol. 114, no.~3, pp.
  462--467, 2017.

\bibitem{liu2015branch}
M.~Liu, Z.~Luo, and A.~Lim, ``A branch-and-cut algorithm for a realistic
  dial-a-ride problem,'' \emph{Transportation Research Part B: Methodological},
  vol.~81, pp. 267--288, 2015.

\bibitem{qu2015branch}
Y.~Qu and J.~F. Bard, ``A branch-and-price-and-cut algorithm for heterogeneous
  pickup and delivery problems with configurable vehicle capacity,''
  \emph{Transportation Science}, vol.~49, no.~2, pp. 254--270, 2015.

\bibitem{parragh2015dial}
S.~N. Parragh, J.~Pinho~de Sousa, and B.~Almada-Lobo, ``The dial-a-ride problem
  with split requests and profits,'' \emph{Transportation Science}, vol.~49,
  no.~2, pp. 311--334, 2015.

\bibitem{parragh2012models}
S.~N. Parragh, J.-F. Cordeau, K.~F. Doerner, and R.~F. Hartl, ``Models and
  algorithms for the heterogeneous dial-a-ride problem with driver-related
  constraints,'' \emph{OR spectrum}, vol.~34, no.~3, pp. 593--633, 2012.

\bibitem{hame2015maximum}
L.~H{\"a}me and H.~Hakula, ``A maximum cluster algorithm for checking the
  feasibility of dial-a-ride instances,'' \emph{Transportation Science},
  vol.~49, no.~2, pp. 295--310, 2015.

\bibitem{cortes2010pickup}
C.~E. Cort{\'e}s, M.~Matamala, and C.~Contardo, ``The pickup and delivery
  problem with transfers: Formulation and a branch-and-cut solution method,''
  \emph{European Journal of Operational Research}, vol. 200, no.~3, pp.
  711--724, 2010.

\bibitem{ordonez2017dynamic}
F.~Ord{\'o}{\~n}ez and M.~M. Dessouky, ``Dynamic ridesharing,'' in
  \emph{Leading developments from INFORMS communities}.\hskip 1em plus 0.5em
  minus 0.4em\relax INFORMS, 2017, pp. 212--236.

\bibitem{molenbruch2017typology}
Y.~Molenbruch, K.~Braekers, and A.~Caris, ``Typology and literature review for
  dial-a-ride problems,'' \emph{Annals of Operations Research}, vol. 259,
  no.~1, pp. 295--325, 2017.

\bibitem{luo2011online}
Y.~Luo and P.~Schonfeld, ``Online rejected-reinsertion heuristics for dynamic
  multivehicle dial-a-ride problem,'' \emph{Transportation research record},
  vol. 2218, no.~1, pp. 59--67, 2011.

\bibitem{chassaing2016els}
M.~Chassaing, C.~Duhamel, and P.~Lacomme, ``An els-based approach with dynamic
  probabilities management in local search for the dial-a-ride problem,''
  \emph{Engineering Applications of Artificial Intelligence}, vol.~48, pp.
  119--133, 2016.

\bibitem{ritzinger2016dynamic}
U.~Ritzinger, J.~Puchinger, and R.~F. Hartl, ``Dynamic programming based
  metaheuristics for the dial-a-ride problem,'' \emph{Annals of Operations
  Research}, vol. 236, no.~2, pp. 341--358, 2016.

\bibitem{dutta2018hashing}
C.~Dutta, ``When hashing met matching: Efficient spatio-temporal search for
  ridesharing,'' in \emph{Proc.~of AAAI}, 2021.

\bibitem{SHARIFAZADEH2022128}
S.~{Sharif Azadeh}, B.~Atasoy, M.~E. Ben-Akiva, M.~Bierlaire, and M.~Maknoon,
  ``Choice-driven dial-a-ride problem for demand responsive mobility service,''
  \emph{Transportation Research Part B: Methodological}, vol. 161, pp.
  128--149, 2022.

\bibitem{de2022influence}
A.~de~Palma, P.~Stokkink, and N.~Geroliminis, ``Influence of dynamic congestion
  with scheduling preferences on carpooling matching with heterogeneous
  users,'' \emph{Transportation Research Part B: Methodological}, vol. 155, pp.
  479--498, 2022.

\bibitem{zhou2022scalable}
Z.~Zhou and C.~Roncoli, ``A scalable vehicle assignment and routing strategy
  for real-time on-demand ridesharing considering endogenous congestion,''
  \emph{Transportation Research Part C: Emerging Technologies}, vol. 139, p.
  103658, 2022.

\bibitem{peng2022investigating}
W.~Peng and L.~Du, ``Investigating optimal carpool scheme by a semi-centralized
  ride-matching approach,'' \emph{IEEE Transactions on Intelligent
  Transportation Systems}, to appear, doi: 10.1109/TITS.2021.3135648.

\bibitem{nguyen2018policy}
D.~T. Nguyen, A.~Kumar, and H.~C. Lau, ``Policy gradient with value function
  approximation for collective multiagent planning,'' in \emph{Proc.~of
  NeurIPS}, 2017.

\bibitem{al2019deeppool}
A.~O. Al-Abbasi, A.~Ghosh, and V.~Aggarwal, ``Deeppool: Distributed model-free
  algorithm for ride-sharing using deep reinforcement learning,'' \emph{IEEE
  Transactions on Intelligent Transportation Systems}, vol.~20, no.~12, pp.
  4714--4727, 2019.

\bibitem{shi2021learning}
D.~Shi, Y.~Tong, Z.~Zhou, B.~Song, W.~Lv, and Q.~Yang, ``Learning to assign:
  Towards fair task assignment in large-scale ride hailing,'' in \emph{Proc.~of
  ACM SIGKDD}, 2021.

\bibitem{oda2021equilibrium}
T.~Oda, ``Equilibrium inverse reinforcement learning for ride-hailing vehicle
  network,'' in \emph{Proc.~of WWW}, 2021.

\bibitem{haliem2021distributed}
M.~Haliem, G.~Mani, V.~Aggarwal, and B.~Bhargava, ``A distributed model-free
  ride-sharing approach for joint matching, pricing, and dispatching using deep
  reinforcement learning,'' \emph{IEEE Transactions on Intelligent
  Transportation Systems}, vol.~22, no.~12, pp. 7931--7942, 2021.

\bibitem{agatz2012optimization}
N.~Agatz, A.~Erera, M.~Savelsbergh, and X.~Wang, ``Optimization for dynamic
  ride-sharing: A review,'' \emph{European Journal of Operational Research},
  vol. 223, no.~2, pp. 295--303, 2012.

\bibitem{furuhata2013ridesharing}
M.~Furuhata, M.~Dessouky, F.~Ord{\'o}{\~n}ez, M.-E. Brunet, X.~Wang, and
  S.~Koenig, ``Ridesharing: The state-of-the-art and future directions,''
  \emph{Transportation Research Part B: Methodological}, vol.~57, pp. 28--46,
  2013.

\bibitem{mourad2019survey}
A.~Mourad, J.~Puchinger, and C.~Chu, ``A survey of models and algorithms for
  optimizing shared mobility,'' \emph{Transportation Research Part B:
  Methodological}, vol. 123, pp. 323--346, 2019.

\bibitem{yan2020dynamic}
C.~Yan, H.~Zhu, N.~Korolko, and D.~Woodard, ``Dynamic pricing and matching in
  ride-hailing platforms,'' \emph{Naval Research Logistics (NRL)}, vol.~67,
  no.~8, pp. 705--724, 2020.

\bibitem{martins2021optimizing}
L.~d.~C. Martins, R.~de~la Torre, C.~G. Corlu, A.~A. Juan, and M.~A. Masmoudi,
  ``Optimizing ride-sharing operations in smart sustainable cities: Challenges
  and the need for agile algorithms,'' \emph{Computers \& Industrial
  Engineering}, vol. 153, p. 107080, 2021.

\bibitem{lin2018efficient}
K.~Lin, R.~Zhao, Z.~Xu, and J.~Zhou, ``Efficient large-scale fleet management
  via multi-agent deep reinforcement learning,'' in \emph{Proc.~of ACM SIGKDD},
  2018.

\bibitem{tang2019deep}
X.~Tang, Z.~Qin, F.~Zhang, Z.~Wang, Z.~Xu, Y.~Ma, H.~Zhu, and J.~Ye, ``A deep
  value-network based approach for multi-driver order dispatching,'' in
  \emph{Proc.~of ACM SIGKDD}, 2019.

\bibitem{veres2019deep}
M.~Veres and M.~Moussa, ``Deep learning for intelligent transportation systems:
  A survey of emerging trends,'' \emph{IEEE Transactions on Intelligent
  transportation systems}, vol.~21, no.~8, pp. 3152--3168, 2019.

\bibitem{farazi2020deep}
N.~P. Farazi, T.~Ahamed, L.~Barua, and B.~Zou, ``Deep reinforcement learning
  and transportation research: A comprehensive review,'' \emph{Transportation
  Research Interdisciplinary Perspectives}, vol.~11, no.~0, 2021.

\bibitem{haghighat2020applications}
A.~K. Haghighat, V.~Ravichandra-Mouli, P.~Chakraborty, Y.~Esfandiari, S.~Arabi,
  and A.~Sharma, ``Applications of deep learning in intelligent transportation
  systems,'' \emph{Journal of Big Data Analytics in Transportation}, vol.~2,
  no.~2, pp. 115--145, 2020.

\bibitem{chakraborty2020review}
J.~Chakraborty, D.~Pandit, F.~Chan, and J.~Xia, ``A review of ride-matching
  strategies for ridesourcing and other similar services,'' \emph{Transport
  Reviews}, pp. 1--22, 2020.

\bibitem{qin2021reinforcement}
Z.~T. Qin, H.~Zhu, and J.~Ye, ``Reinforcement learning for ridesharing: A
  survey,'' in \emph{Proc.~of IEEE ITSC}, 2021.

\bibitem{qin2022reinforcement}
------, ``Reinforcement learning for ridesharing: An extended survey,''
  \emph{Transportation Research Part C: Emerging Technologies}, vol. 144, p.
  103852, 2022.

\bibitem{dickerson2018allocation}
J.~Dickerson, K.~Sankararaman, A.~Srinivasan, and P.~Xu, ``Allocation problems
  in ride-sharing platforms: Online matching with offline reusable resources,''
  in \emph{Proc.~of AAAI}, 2018.

\bibitem{xu2018large}
Z.~Xu, Z.~Li, Q.~Guan, D.~Zhang, Q.~Li, J.~Nan, C.~Liu, W.~Bian, and J.~Ye,
  ``Large-scale order dispatch in on-demand ride-hailing platforms: A learning
  and planning approach,'' in \emph{Proc.~of ACM SIGKDD}, 2018.

\bibitem{alonso2017predictive}
J.~Alonso-Mora, A.~Wallar, and D.~Rus, ``Predictive routing for autonomous
  mobility-on-demand systems with ride-sharing,'' in \emph{Proc.~of IEEE/RSJ
  IROS}, 2017.

\bibitem{liu2019globally}
Y.~Liu, W.~Skinner, and C.~Xiang, ``Globally-optimized realtime supply-demand
  matching in on-demand ridesharing,'' in \emph{Proc.~of WWW}, 2019.

\bibitem{lin2019probabilistic}
Q.~Lin, W.~Xu, M.~Chen, and X.~Lin, ``A probabilistic approach for demand-aware
  ride-sharing optimization,'' in \emph{Proc.~of ACM Mobihoc}, 2019.

\bibitem{wang2019data}
C.~Wang, Y.~Hou, and M.~Barth, ``Data-driven multi-step demand prediction for
  ride-hailing services using convolutional neural network,'' in \emph{Proc.~of
  SAI}, 2019.

\bibitem{jindal2018optimizing}
I.~Jindal, Z.~T. Qin, X.~Chen, M.~Nokleby, and J.~Ye, ``Optimizing taxi carpool
  policies via reinforcement learning and spatio-temporal mining,'' in
  \emph{Proc.~of IEEE BigData}, 2018.

\bibitem{zhou2019multi}
M.~Zhou, J.~Jin, W.~Zhang, Z.~Qin, Y.~Jiao, C.~Wang, G.~Wu, Y.~Yu, and J.~Ye,
  ``Multi-agent reinforcement learning for order-dispatching via order-vehicle
  distribution matching,'' in \emph{Proc.~of ACM CIKM}, 2019.

\bibitem{liu2022personalized}
S.~Liu and H.~Jiang, ``Personalized route recommendation for ride-hailing with
  deep inverse reinforcement learning and real-time traffic conditions,''
  \emph{Transportation Research Part E: Logistics and Transportation Review},
  vol. 164, p. 102780, 2022.

\bibitem{shah2020sride}
I.~Shah, M.~El~Affendi, and B.~Qureshi, ``Sride: An online system for multi-hop
  ridesharing,'' \emph{Sustainability}, vol.~12, no.~22, p. 9633, 2020.

\bibitem{xu2020highly}
Y.~Xu, L.~Kulik, R.~Borovica-Gajic, A.~Aldwyish, and J.~Qi, ``Highly efficient
  and scalable multi-hop ride-sharing,'' in \emph{Proc.~of ACM SIGSPATIAL},
  2020.

\bibitem{li2020trip}
S.~Li, M.~Li, and V.~C. Lee, ``Trip-vehicle assignment algorithms for
  ride-sharing,'' in \emph{Proc.~of COCOA}, 2020.

\bibitem{xu2020recommender}
Z.~Xu, C.~Men, P.~Li, B.~Jin, G.~Li, Y.~Yang, C.~Liu, B.~Wang, and X.~Qie,
  ``When recommender systems meet fleet management: Practical study in online
  driver repositioning system,'' in \emph{Proc.~of WWW}, 2020.

\bibitem{chaudhari2020learn}
H.~A. Chaudhari, J.~W. Byers, and E.~Terzi, ``Learn to earn: Enabling
  coordination within a ride hailing fleet,'' in \emph{Proc.~of IEEE BigData},
  2020.

\bibitem{gao2020learning}
J.~Gao, X.~Li, C.~Wang, and X.~Huang, ``Learning-based open driver guidance and
  rebalancing for reducing riders’ wait time in ride-hailing platforms,'' in
  \emph{Proc.~of IEEE ISC2}, 2020.

\bibitem{zhang2020multiple}
Z.~Zhang, Y.~Li, and H.~Dong, ``Multiple-feature-based vehicle supply--demand
  difference prediction method for social transportation,'' \emph{IEEE
  Transactions on Computational Social Systems}, vol.~7, no.~4, pp. 1095--1103,
  2020.

\bibitem{guo2021multi}
Y.~Guo, Y.~Zhang, Y.~Boulaksil, and N.~Tian, ``Multi-dimensional spatiotemporal
  demand forecasting and service vehicle dispatching for online car-hailing
  platforms,'' \emph{International Journal of Production Research}, pp. 1--22,
  2021.

\bibitem{zhang2021mlrnn}
C.~Zhang, F.~Zhu, Y.~Lv, P.~Ye, and F.-Y. Wang, ``Mlrnn: Taxi demand prediction
  based on multi-level deep learning and regional heterogeneity analysis,''
  \emph{IEEE Transactions on Intelligent Transportation Systems}, 2021, to
  appear, doi: 10.1109/TITS.2021.3080511.

\bibitem{ke2021joint}
J.~Ke, S.~Feng, Z.~Zhu, H.~Yang, and J.~Ye, ``Joint predictions of multi-modal
  ride-hailing demands: A deep multi-task multi-graph learning-based
  approach,'' \emph{Transportation Research Part C: Emerging Technologies},
  vol. 127, p. 103063, 2021.

\bibitem{chen2022h}
Z.~Chen, K.~Liu, J.~Wang, and T.~Yamamoto, ``H-convlstm-based bagging learning
  approach for ride-hailing demand prediction considering imbalance problems
  and sparse uncertainty,'' \emph{Transportation Research Part C: Emerging
  Technologies}, vol. 140, p. 103709, 2022.

\bibitem{o2021using}
K.~O’Keeffe, S.~Anklesaria, P.~Santi, and C.~Ratti, ``Using reinforcement
  learning to minimize taxi idle times,'' \emph{Journal of Intelligent
  Transportation Systems}, pp. 1--16, 2021.

\bibitem{ji2020spatio}
S.~Ji, Z.~Wang, T.~Li, and Y.~Zheng, ``Spatio-temporal feature fusion for
  dynamic taxi route recommendation via deep reinforcement learning,''
  \emph{Knowledge-Based Systems}, vol. 205, p. 106302, 2020.

\bibitem{oda2018distributed}
T.~Oda and Y.~Tachibana, ``Distributed fleet control with maximum entropy deep
  reinforcement learning,'' in \emph{Proc.~of NeurIPS}, 2018.

\bibitem{liu2020context}
Z.~Liu, J.~Li, and K.~Wu, ``Context-aware taxi dispatching at city-scale using
  deep reinforcement learning,'' \emph{IEEE Transactions on Intelligent
  Transportation Systems}, vol.~23, no.~3, pp. 1996--2009, 2022.

\bibitem{oda2018movi}
T.~Oda and C.~Joe-Wong, ``Movi: A model-free approach to dynamic fleet
  management,'' in \emph{Proc.~of IEEE INFOCOM}, 2018.

\bibitem{sadeghiyengejeh2021re}
A.~Sadeghiyengejeh and S.~L. Smith, ``Re-balancing self-interested drivers in
  ride-sharing networks to improve customer wait-time,'' \emph{IEEE
  Transactions on Control of Network Systems}, 2021.

\bibitem{he2019spatio}
S.~He and K.~G. Shin, ``Spatio-temporal capsule-based reinforcement learning
  for mobility-on-demand network coordination,'' in \emph{Proc.~of WWW}, 2019.

\bibitem{singh2021distributed}
A.~Singh, A.~O. Al-Abbasi, and V.~Aggarwal, ``A distributed model-free
  algorithm for multi-hop ride-sharing using deep reinforcement learning,''
  \emph{IEEE Transactions on Intelligent Transportation Systems}, 2021, to
  appear, doi: 10.1109/TITS.2021.3083740.

\bibitem{chau2020decentralized}
S.~C.-K. Chau, S.~Shen, and Y.~Zhou, ``Decentralized ride-sharing and
  vehicle-pooling based on fair cost-sharing mechanisms,'' \emph{IEEE
  Transactions on Intelligent Transportation Systems}, vol.~23, no.~3, pp.
  1936--1946, 2022.

\bibitem{chen2020order}
Z.~Chen, P.~Li, J.~Xiao, L.~Nie, and Y.~Liu, ``An order dispatch system based
  on reinforcement learning for ride sharing services,'' in \emph{Proc.~of IEEE
  HPCC/SmartCity/DSS}, 2020.

\bibitem{agussurja2019state}
L.~Agussurja, S.-F. Cheng, and H.~C. Lau, ``A state aggregation approach for
  stochastic multiperiod last-mile ride-sharing problems,''
  \emph{Transportation Science}, vol.~53, no.~1, pp. 148--166, 2019.

\bibitem{kullman2022dynamic}
N.~D. Kullman, M.~Cousineau, J.~C. Goodson, and J.~E. Mendoza, ``Dynamic
  ride-hailing with electric vehicles,'' \emph{Transportation Science},
  vol.~56, no.~3, pp. 775--794, 2022.

\bibitem{wang2023optimization}
D.~Wang, Q.~Wang, Y.~Yin, and T.~Cheng, ``Optimization of ride-sharing with
  passenger transfer via deep reinforcement learning,'' \emph{Transportation
  Research Part E: Logistics and Transportation Review}, vol. 172, p. 103080,
  2023.

\bibitem{wang2019adaptive}
Y.~Wang, Y.~Tong, C.~Long, P.~Xu, K.~Xu, and W.~Lv, ``Adaptive dynamic
  bipartite graph matching: A reinforcement learning approach,'' in
  \emph{Proc.~of IEEE ICDE}, 2019.

\bibitem{qin2021optimizing}
G.~Qin, Q.~Luo, Y.~Yin, J.~Sun, and J.~Ye, ``Optimizing matching time intervals
  for ride-hailing services using reinforcement learning,''
  \emph{Transportation Research Part C: Emerging Technologies}, vol. 129, p.
  103239, 2021.

\bibitem{hong2017commuter}
Z.~Hong, Y.~Chen, H.~S. Mahmassani, and S.~Xu, ``Commuter ride-sharing using
  topology-based vehicle trajectory clustering: Methodology, application and
  impact evaluation,'' \emph{Transportation Research Part C: Emerging
  Technologies}, vol.~85, pp. 573--590, 2017.

\bibitem{parimala2011survey}
M.~Parimala, D.~Lopez, and N.~Senthilkumar, ``A survey on density based
  clustering algorithms for mining large spatial databases,''
  \emph{International Journal of Advanced Science and Technology}, vol.~31,
  no.~1, pp. 59--66, 2011.

\bibitem{zheng2018order}
L.~Zheng, L.~Chen, and J.~Ye, ``Order dispatch in price-aware ridesharing,''
  \emph{Proceedings of the VLDB Endowment}, vol.~11, no.~8, pp. 853--865, 2018.

\bibitem{shen2019roo}
B.~Shen, B.~Cao, Y.~Zhao, H.~Zuo, W.~Zheng, and Y.~Huang, ``Roo: Route planning
  algorithm for ride sharing systems on large-scale road networks,'' in
  \emph{Proc.~of IEEE BigComp}, 2019.

\bibitem{trasarti2011mining}
R.~Trasarti, F.~Pinelli, M.~Nanni, and F.~Giannotti, ``Mining mobility user
  profiles for car pooling,'' in \emph{Proc.~of ACM SIGKDD}, 2011.

\bibitem{mitropoulos2021systematic}
L.~Mitropoulos, A.~Kortsari, and G.~Ayfantopoulou, ``A systematic literature
  review of ride-sharing platforms, user factors and barriers,'' \emph{European
  Transport Research Review}, vol.~13, pp. 1--22, 2021.

\bibitem{yatnalkar2020enhanced}
G.~Yatnalkar, H.~S. Narman, and H.~Malik, ``An enhanced ride sharing model
  based on human characteristics and machine learning recommender system,''
  \emph{Procedia Computer Science}, vol. 170, pp. 626--633, 2020.

\bibitem{narman2021enhanced}
H.~S. Narman, H.~Malik, and G.~Yatnalkar, ``An enhanced ride sharing model
  based on human characteristics, machine learning recommender system, and user
  threshold time,'' \emph{Journal of Ambient Intelligence and Humanized
  Computing}, vol.~12, no.~1, pp. 13--26, 2021.

\bibitem{levinger2020human}
C.~Levinger, N.~Hazon, and A.~Azaria, ``Human satisfaction as the ultimate goal
  in ridesharing,'' \emph{Future Generation Computer Systems}, vol. 112, pp.
  176--184, 2020.

\bibitem{montazery2016learning}
M.~Montazery and N.~Wilson, ``Learning user preferences in matching for
  ridesharing.'' in \emph{Proc.~of ICAART}, 2016.

\bibitem{montazery2018new}
------, ``A new approach for learning user preferences for a ridesharing
  application,'' in \emph{Transactions on Computational Collective Intelligence
  XXVIII}.\hskip 1em plus 0.5em minus 0.4em\relax Springer, 2018, pp. 1--24.

\bibitem{tang2020efficient}
L.~Tang, Z.~Liu, Y.~Zhao, Z.~Duan, and J.~Jia, ``Efficient ridesharing
  framework for ride-matching via heterogeneous network embedding,'' \emph{ACM
  Transactions on Knowledge Discovery from Data (TKDD)}, vol.~14, no.~3, pp.
  1--24, 2020.

\bibitem{sun2012mining}
Y.~Sun and J.~Han, ``Mining heterogeneous information networks: principles and
  methodologies,'' \emph{Synthesis Lectures on Data Mining and Knowledge
  Discovery}, vol.~3, no.~2, pp. 1--159, 2012.

\bibitem{mikolov2013efficient}
T.~Mikolov, K.~Chen, G.~Corrado, and J.~Dean, ``Efficient estimation of word
  representations in vector space,'' in \emph{Proc.~of ICLR Workshop}, 2013.

\bibitem{dong2017metapath2vec}
Y.~Dong, N.~V. Chawla, and A.~Swami, ``metapath2vec: Scalable representation
  learning for heterogeneous networks,'' in \emph{Proc.~of ACM SIGKDD}, 2017.

\bibitem{zhang2017taxi}
L.~Zhang, T.~Hu, Y.~Min, G.~Wu, J.~Zhang, P.~Feng, P.~Gong, and J.~Ye, ``A taxi
  order dispatch model based on combinatorial optimization,'' in \emph{Proc.~of
  ACM SIGKDD}, 2017.

\bibitem{friedman2001elements}
J.~Friedman, T.~Hastie, R.~Tibshirani \emph{et~al.}, \emph{The elements of
  statistical learning}.\hskip 1em plus 0.5em minus 0.4em\relax Springer series
  in statistics New York, 2001, vol.~1, no.~10.

\bibitem{mason1999boosting}
L.~Mason, J.~Baxter, P.~Bartlett, and M.~Frean, ``Boosting algorithms as
  gradient descent in function space,'' in \emph{Proc.~of NeurIPS}, 1999.

\bibitem{schleibaum2020human}
S.~Schleibaum and J.~P. M{\"u}ller, ``Human-centric ridesharing on large scale
  by explaining ai-generated assignments,'' in \emph{Proc.~of EAI GOODTECHS},
  2020.

\bibitem{jonker1986improving}
R.~Jonker and T.~Volgenant, ``Improving the hungarian assignment algorithm,''
  \emph{Operations Research Letters}, vol.~5, no.~4, pp. 171--175, 1986.

\bibitem{guo2020spatiotemporal}
Y.~Guo, Y.~Zhang, J.~Yu, and X.~Shen, ``A spatiotemporal thermo guidance based
  real-time online ride-hailing dispatch framework,'' \emph{IEEE Access},
  vol.~8, pp. 115\,063--115\,077, 2020.

\bibitem{breiman2001random}
L.~Breiman, ``Random forests,'' \emph{Machine learning}, vol.~45, no.~1, pp.
  5--32, 2001.

\bibitem{guo2020deep}
G.~Guo and Y.~Xu, ``A deep reinforcement learning approach to ride-sharing
  vehicles dispatching in autonomous mobility-on-demand systems,'' \emph{IEEE
  Intelligent Transportation Systems Magazine}, 2020.

\bibitem{van2016deep}
H.~Van~Hasselt, A.~Guez, and D.~Silver, ``Deep reinforcement learning with
  double q-learning,'' in \emph{Proc.~of AAAI}, 2016.

\bibitem{sutton1999reinforcement}
R.~S. Sutton and A.~G. Barto, ``Reinforcement learning,'' \emph{Journal of
  Cognitive Neuroscience}, vol.~11, no.~1, pp. 126--134, 1999.

\bibitem{gueriau2018samod}
M.~Gu{\'e}riau and I.~Dusparic, ``Samod: Shared autonomous mobility-on-demand
  using decentralized reinforcement learning,'' in \emph{Proc.~of IEEE ITSC},
  2018.

\bibitem{gueriau2020shared}
M.~Gu{\'e}riau, F.~Cugurullo, R.~A. Acheampong, and I.~Dusparic, ``Shared
  autonomous mobility on demand: A learning-based approach and its performance
  in the presence of traffic congestion,'' \emph{IEEE Intelligent
  Transportation Systems Magazine}, vol.~12, no.~4, pp. 208--218, 2020.

\bibitem{wang2018deep}
Z.~Wang, Z.~Qin, X.~Tang, J.~Ye, and H.~Zhu, ``Deep reinforcement learning with
  knowledge transfer for online rides order dispatching,'' in \emph{Proc.~of
  IEEE ICDM}, 2018.

\bibitem{mnih2015human}
V.~Mnih, K.~Kavukcuoglu, D.~Silver, A.~A. Rusu, J.~Veness, M.~G. Bellemare,
  A.~Graves, M.~Riedmiller, A.~K. Fidjeland, G.~Ostrovski \emph{et~al.},
  ``Human-level control through deep reinforcement learning,'' \emph{nature},
  vol. 518, no. 7540, pp. 529--533, 2015.

\bibitem{de2020efficient}
O.~De~Lima, H.~Shah, T.-S. Chu, and B.~Fogelson, ``Efficient ridesharing
  dispatch using multi-agent reinforcement learning,'' \emph{arXiv preprint
  arXiv:2006.10897}, 2020.

\bibitem{rashid2018qmix}
T.~Rashid, M.~Samvelyan, C.~Schroeder, G.~Farquhar, J.~Foerster, and
  S.~Whiteson, ``Qmix: Monotonic value function factorisation for deep
  multi-agent reinforcement learning,'' in \emph{Proc.~of ICML}, 2018.

\bibitem{li2019efficient}
M.~Li, Z.~Qin, Y.~Jiao, Y.~Yang, J.~Wang, C.~Wang, G.~Wu, and J.~Ye,
  ``Efficient ridesharing order dispatching with mean field multi-agent
  reinforcement learning,'' in \emph{Proc.~of WWW}, 2019.

\bibitem{bicocchi2014investigating}
N.~Bicocchi and M.~Mamei, ``Investigating ride sharing opportunities through
  mobility data analysis,'' \emph{Pervasive and Mobile Computing}, vol.~14, pp.
  83--94, 2014.

\bibitem{blei2003latent}
D.~M. Blei, A.~Y. Ng, and M.~I. Jordan, ``Latent dirichlet allocation,''
  \emph{the Journal of machine Learning research}, vol.~3, pp. 993--1022, 2003.

\bibitem{lasmar2019rsrs}
E.~L. Lasmar, F.~O. de~Paula, R.~L. Rosa, J.~I. Abrah{\~a}o, and D.~Z.
  Rodr{\'\i}guez, ``Rsrs: Ridesharing recommendation system based on social
  networks to improve the user’s qoe,'' \emph{IEEE Transactions on
  Intelligent Transportation Systems}, vol.~20, no.~12, pp. 4728--4740, 2019.

\bibitem{liang2020mobility}
Y.~Liang, Z.~Ding, T.~Ding, and W.-J. Lee, ``Mobility-aware charging scheduling
  for shared on-demand electric vehicle fleet using deep reinforcement
  learning,'' \emph{IEEE Transactions on Smart Grid}, vol.~12, no.~2, pp.
  1380--1393, 2020.

\bibitem{tang2020online}
X.~Tang, M.~Li, X.~Lin, and F.~He, ``Online operations of automated electric
  taxi fleets: An advisor-student reinforcement learning framework,''
  \emph{Transportation Research Part C: Emerging Technologies}, vol. 121, p.
  102844, 2020.

\bibitem{liang2021integrated}
E.~Liang, K.~Wen, W.~H. Lam, A.~Sumalee, and R.~Zhong, ``An integrated
  reinforcement learning and centralized programming approach for online taxi
  dispatching,'' \emph{IEEE Transactions on Neural Networks and Learning
  Systems}, 2021.

\bibitem{fluri2019learning}
C.~Fluri, C.~Ruch, J.~Zilly, J.~Hakenberg, and E.~Frazzoli, ``Learning to
  operate a fleet of cars,'' in \emph{Proc.~of IEEE ITSC}, 2019.

\bibitem{lloyd1982least}
S.~Lloyd, ``Least squares quantization in pcm,'' \emph{IEEE transactions on
  information theory}, vol.~28, no.~2, pp. 129--137, 1982.

\bibitem{deng2020multi}
Y.~Deng, H.~Chen, S.~Shao, J.~Tang, J.~Pi, and A.~Gupta, ``Multi-objective
  vehicle rebalancing for ridehailing system using a reinforcement learning
  approach,'' \emph{Journal of Management Science and Engineering}, 2021.

\bibitem{schulman2017proximal}
J.~Schulman, F.~Wolski, P.~Dhariwal, A.~Radford, and O.~Klimov, ``Proximal
  policy optimization algorithms,'' \emph{arXiv preprint arXiv:1707.06347},
  2017.

\bibitem{shi2019optimal}
D.~Shi, X.~Li, M.~Li, J.~Wang, P.~Li, and M.~Pan, ``Optimal transportation
  network company vehicle dispatching via deep deterministic policy gradient,''
  in \emph{Proc.~of WASA}, 2019.

\bibitem{silver2014deterministic}
D.~Silver, G.~Lever, N.~Heess, T.~Degris, D.~Wierstra, and M.~Riedmiller,
  ``Deterministic policy gradient algorithms,'' in \emph{Proc.~of ICML}, 2014.

\bibitem{shou2020reward}
Z.~Shou and X.~Di, ``Reward design for driver repositioning using multi-agent
  reinforcement learning,'' \emph{Transportation research part C: emerging
  technologies}, vol. 119, p. 102738, 2020.

\bibitem{riley2020real}
C.~Riley, P.~van Hentenryck, and E.~Yuan, ``Real-time dispatching of
  large-scale ride-sharing systems: Integrating optimization, machine learning,
  and model predictive control,'' in \emph{Proc.~of IJCAI}, 2020.

\bibitem{iglesias2018data}
R.~Iglesias, F.~Rossi, K.~Wang, D.~Hallac, J.~Leskovec, and M.~Pavone,
  ``Data-driven model predictive control of autonomous mobility-on-demand
  systems,'' in \emph{Proc.~of IEEE ICRA}, 2018.

\bibitem{xu2018taxi}
J.~Xu, R.~Rahmatizadeh, L.~B{\"o}l{\"o}ni, and D.~Turgut, ``Taxi dispatch
  planning via demand and destination modeling,'' in \emph{Proc.~of IEEE LCN},
  2018.

\bibitem{cheng18taxis}
S.-F. Cheng, S.~S. Jha, and R.~Rajendram, ``Taxis strike back: A field trial of
  the driver guidance system,'' in \emph{Proc.~of AAMAS}, 2018.

\bibitem{li2020data}
X.~Li, C.~Wang, X.~Huang, and Y.~Nie, ``A data-driven dynamic stochastic
  programming framework for ride-sharing rebalancing problem under demand
  uncertainty,'' in \emph{Proc.~of IEEE ISPA-BDCloud-SocialCom-SustainCom},
  2020.

\bibitem{pouls2020idle}
M.~Pouls, A.~Meyer, and N.~Ahuja, ``Idle vehicle repositioning for dynamic
  ride-sharing,'' in \emph{Proc.~of ICCL}, 2020.

\bibitem{gurobi}
``Gurobi,'' \url{https://www.gurobi.com}, visted on May 25, 2022.

\bibitem{chen2016xgboost}
T.~Chen and C.~Guestrin, ``Xgboost: A scalable tree boosting system,'' in
  \emph{Proc.~of ACM SIGKDD}, 2016.

\bibitem{makhorin2008glpk}
A.~Makhorin, ``Glpk (gnu linear programming kit),'' \emph{http://www. gnu.
  org/s/glpk/glpk. html}, 2008.

\bibitem{ke2019optimizing}
J.~Ke, F.~Xiao, H.~Yang, and J.~Ye, ``Optimizing online matching for
  ride-sourcing services with multi-agent deep reinforcement learning,''
  \emph{CoRR}, vol. abs/1902.06228, 2019.

\bibitem{castagna2020demand}
A.~Castagna, M.~Gu{\'e}riau, G.~Vizzari, and I.~Dusparic, ``Demand-responsive
  zone generation for real-time vehicle rebalancing in ride-sharing fleets,''
  in \emph{Proc.~of ATT}, 2020.

\bibitem{castagna2021demand}
------, ``Demand-responsive rebalancing zone generation for reinforcement
  learning-based on-demand mobility,'' \emph{AI Communications}, pp. 1--16,
  2021.

\bibitem{tang2021value}
X.~Tang, F.~Zhang, Y.~Wang, D.~Shi, B.~Song, Y.~Tong, H.~Zhu, J.~Ye
  \emph{et~al.}, ``Value function is all you need: A unified learning framework
  for ride hailing platforms,'' in \emph{Proc.~of ACM SIGKDD}, 2021.

\bibitem{verma2017augmenting}
T.~Verma, P.~Varakantham, S.~Kraus, and H.~C. Lau, ``Augmenting decisions of
  taxi drivers through reinforcement learning for improving revenues,'' in
  \emph{Proc.~of ICAPS}, 2017.

\bibitem{provoostdemandprop}
``Demandprop: a scalable algorithm for real-time predictive positioning of
  fleets in dynamic ridesharing systems,''
  \url{https://www.itrl.kth.se/polopoly_fs/1.1060792.1616668658!/PREDICT_preliminary_paper_2503.pdf},
  visted on May 25, 2022.

\bibitem{jiao20deep}
Y.~Jiao, X.~Tang, Z.~T. Qin, S.~Li, F.~Zhang, H.~Zhu, and J.~Ye, ``A deep
  value-based policy search approach for real-world vehicle repositioning on
  mobility-on-demand platforms,'' in \emph{Proc.~of NeurIPS Deep RL Workshop},
  2020.

\bibitem{jiao2021real}
Y.~Jiao, X.~Tang, Z.~Qin, S.~Li, F.~Zhang, H.~Zhu, and J.~Ye, ``Real-world
  ride-hailing vehicle repositioning using deep reinforcement learning,''
  \emph{Transportation Research Part C: Emerging Technologies}, vol. 130, p.
  103289, 2021.

\bibitem{kim2020optimizing}
J.~Kim and K.~Kim, ``Optimizing large-scale fleet management on a road network
  using multi-agent deep reinforcement learning with graph neural network,'' in
  \emph{Proc.~of IEEE ITSC}, 2021.

\bibitem{verma2019entropy}
T.~Verma, P.~Varakantham, and H.~C. Lau, ``Entropy based independent learning
  in anonymous multi-agent settings,'' in \emph{Proc.~of ICAPS}, 2019.

\bibitem{jaynes1957information}
E.~T. Jaynes, ``Information theory and statistical mechanics,'' \emph{Physical
  review}, vol. 106, no.~4, p. 620, 1957.

\bibitem{kemker2018measuring}
R.~Kemker, M.~McClure, A.~Abitino, T.~Hayes, and C.~Kanan, ``Measuring
  catastrophic forgetting in neural networks,'' in \emph{Proc.~of AAAI}, 2018.

\bibitem{haliem2020adapool}
M.~Haliem, V.~Aggarwal, and B.~Bhargava, ``Adapool: An adaptive model-free
  ride-sharing approach for dispatching using deep reinforcement learning,'' in
  \emph{Proc.~of ACM BuildSys}, 2020.

\bibitem{haliem2021adapool}
------, ``Adapool: A diurnal-adaptive fleet management framework using
  model-free deep reinforcement learning and change point detection,''
  \emph{IEEE Transactions on Intelligent Transportation Systems}, vol.~23,
  no.~3, pp. 2471--2481, 2022.

\bibitem{lei2019optimal}
Z.~Lei, X.~Qian, and S.~V. Ukkusuri, ``Optimal proactive vehicle relocation for
  on-demand mobility service with deep convolution-lstm network,'' in
  \emph{Proc.~of IEEE ITSC}, 2019.

\bibitem{khetarpal2022towards}
K.~Khetarpal, M.~Riemer, I.~Rish, and D.~Precup, ``Towards continual
  reinforcement learning: A review and perspectives,'' \emph{Journal of
  Artificial Intelligence Research}, vol.~75, pp. 1401--1476, 2022.

\bibitem{xie2021deep}
A.~Xie, J.~Harrison, and C.~Finn, ``Deep reinforcement learning amidst
  continual structured non-stationarity,'' in \emph{Proc.~of ICML}, 2021.

\bibitem{mao2021near}
W.~Mao, K.~Zhang, R.~Zhu, D.~Simchi-Levi, and T.~Basar, ``Near-optimal
  model-free reinforcement learning in non-stationary episodic mdps,'' in
  \emph{Proc.~of ICML}, 2021.

\bibitem{liu2022deep}
Y.~Liu, F.~Wu, C.~Lyu, S.~Li, J.~Ye, and X.~Qu, ``Deep dispatching: A deep
  reinforcement learning approach for vehicle dispatching on online
  ride-hailing platform,'' \emph{Transportation Research Part E: Logistics and
  Transportation Review}, vol. 161, p. 102694, 2022.

\bibitem{he2020spatio}
S.~He and K.~G. Shin, ``Spatio-temporal capsule-based reinforcement learning
  for mobility-on-demand coordination,'' \emph{IEEE Transactions on Knowledge
  and Data Engineering}, 2020.

\bibitem{yu2019markov}
X.~Yu, S.~Gao, X.~Hu, and H.~Park, ``A markov decision process approach to
  vacant taxi routing with e-hailing,'' \emph{Transportation Research Part B:
  Methodological}, vol. 121, pp. 114--134, 2019.

\bibitem{singh2019reinforcement}
A.~Singh, A.~Alabbasi, and V.~Aggarwal, ``A reinforcement learning based
  algorithm for multi-hop ride-sharing: Model-free approach,'' in
  \emph{Proc.~of NeurIPS ML4AD Workshop}, 2019.

\bibitem{haliem2020distributed-a}
M.~Haliem, G.~Mani, V.~Aggarwal, and B.~Bhargava, ``A distributed model-free
  ride-sharing algorithm with pricing using deep reinforcement learning,'' in
  \emph{Proc.~of CSCS}, 2020.

\bibitem{manchella2020passgoodpool}
K.~Manchella, M.~Haliem, V.~Aggarwal, and B.~Bhargava, ``Passgoodpool: Joint
  passengers and goods fleet management with reinforcement learning aided
  pricing, matching, and route planning,'' \emph{IEEE Transactions on
  Intelligent Transportation Systems}, vol.~23, no.~4, pp. 3866--3877, 2022.

\bibitem{manchella2021flexpool}
K.~Manchella, A.~K. Umrawal, and V.~Aggarwal, ``Flexpool: A distributed
  model-free deep reinforcement learning algorithm for joint passengers and
  goods transportation,'' \emph{IEEE Transactions on Intelligent Transportation
  Systems}, vol.~22, no.~4, pp. 2035--2047, 2021.

\bibitem{guo2022deep}
G.~Guo and Y.~Xu, ``A deep reinforcement learning approach to ride-sharing
  vehicle dispatching in autonomous mobility-on-demand systems,'' \emph{IEEE
  Intelligent Transportation Systems Magazine}, vol.~14, no.~1, 2022.

\bibitem{li2020balancing}
J.~Li and V.~H. Allan, ``Balancing taxi distribution in a city-scale dynamic
  ridesharing service: A hybrid solution based on demand learning,'' in
  \emph{Proc.~of ACM SIGKDD}, 2021.

\bibitem{p668-gy46-22}
\BIBentryALTinterwordspacing
U.~Kaggle, ``Uber pickups in new york city,'' 2022. [Online]. Available:
  \url{https://dx.doi.org/10.21227/p668-gy46}
\BIBentrySTDinterwordspacing

\bibitem{tlctrip}
``{New York} city taxi and limousine commission trip record data,''
  \url{https://www1.nyc.gov/site/tlc/about/tlc-trip-record-data.page}, visted
  on May 25, 2022.

\bibitem{gaiadata}
``Didi gaia open dataset,''
  \url{https://outreach.didichuxing.com/app-vue/DataList02}, visted on May 25,
  2022.

\bibitem{chicagoalll}
``Chicago taxi data (2013-to-present),''
  \url{https://data.cityofchicago.org/Transportation/Taxi-Trips/wrvz-psew},
  visted on May 25, 2022.

\bibitem{tdrive}
``T-drive trajectory data,''
  \url{https://www.microsoft.com/en-us/research/publication/t-drive-trajectory-data-sample/},
  visted on May 25, 2022.

\bibitem{yuan2010t}
J.~Yuan, Y.~Zheng, C.~Zhang, W.~Xie, X.~Xie, G.~Sun, and Y.~Huang, ``T-drive:
  driving directions based on taxi trajectories,'' in \emph{Proc.~of
  SIGSPATIAL}, 2010.

\bibitem{yuan2011driving}
J.~Yuan, Y.~Zheng, X.~Xie, and G.~Sun, ``Driving with knowledge from the
  physical world,'' in \emph{Proc.~of ACM SIGKDD}, 2011.

\bibitem{geolifetraj}
``Gps trajectories with transportation mode labels collected in ({Microsoft
  Research Asia}) geolife project,''
  \url{https://www.microsoft.com/en-us/download/details.aspx?id=52367}, visted
  on May 25, 2022.

\bibitem{zheng2008learning}
Y.~Zheng, L.~Liu, L.~Wang, and X.~Xie, ``Learning transportation mode from raw
  gps data for geographic applications on the web,'' in \emph{Proc.~of WWW},
  2008.

\bibitem{zheng2008understanding}
Y.~Zheng, Q.~Li, Y.~Chen, X.~Xie, and W.-Y. Ma, ``Understanding mobility based
  on gps data,'' in \emph{Proc.~of UbiComp}, 2008.

\bibitem{zheng2010understanding}
Y.~Zheng, Y.~Chen, Q.~Li, X.~Xie, and W.-Y. Ma, ``Understanding transportation
  modes based on gps data for web applications,'' \emph{ACM Transactions on the
  Web (TWEB)}, vol.~4, no.~1, pp. 1--36, 2010.

\bibitem{lian2018one}
J.~Lian and L.~Zhang, ``One-month beijing taxi gps trajectory dataset with taxi
  ids and vehicle status,'' in \emph{Proc.~of DATA}, 2018.

\bibitem{portokaggle}
``Porto taxi data (2013-2014),''
  \url{https://www.kaggle.com/crailtap/taxi-trajectory}, visted on May 25,
  2022.

\bibitem{c7j010-22}
\BIBentryALTinterwordspacing
M.~Piorkowski, N.~Sarafijanovic-Djukic, and M.~Grossglauser, ``Crawdad
  epfl/mobility,'' 2022. [Online]. Available:
  \url{https://dx.doi.org/10.15783/C7J010}
\BIBentrySTDinterwordspacing

\bibitem{c7qc7m-22}
\BIBentryALTinterwordspacing
L.~Bracciale, M.~Bonola, P.~Loreti, G.~Bianchi, R.~Amici, and A.~Rabuffi,
  ``Crawdad roma/taxi,'' 2022. [Online]. Available:
  \url{https://dx.doi.org/10.15783/C7QC7M}
\BIBentrySTDinterwordspacing

\bibitem{y8vk-wj40-22}
\BIBentryALTinterwordspacing
A.~Mehmood and F.~Mehmood, ``Vehicular trajectories from jeju, south korea,''
  2022. [Online]. Available: \url{https://dx.doi.org/10.21227/y8vk-wj40}
\BIBentrySTDinterwordspacing

\bibitem{grabsource}
``The source of the {Grab-Posisi} dataset,''
  https://engineering.grab.com/grab-posisi, visted on May 25, 2022.

\bibitem{huang2019grab}
X.~Huang, Y.~Yin, S.~Lim, G.~Wang, B.~Hu, J.~Varadarajan, S.~Zheng, A.~Bulusu,
  and R.~Zimmermann, ``Grab-posisi: An extensive real-life gps trajectory
  dataset in southeast asia,'' in \emph{Proc.~of ACM PredictGIS}, 2019.

\bibitem{2877-mk46-19}
\BIBentryALTinterwordspacing
K.~Liu, ``Demands for bus ridesharing in shanghai, china,'' 2019. [Online].
  Available: \url{https://dx.doi.org/10.21227/2877-mk46}
\BIBentrySTDinterwordspacing

\bibitem{liu2020optimization}
K.~Liu and J.~Liu, ``Optimization approach to improve the ridesharing success
  rate in the bus ridesharing service,'' \emph{IEEE Access}, vol.~8, pp.
  208\,296--208\,310, 2020.

\bibitem{foursquare}
``Foursquare check-in datasets,''
  \url{https://sites.google.com/site/yangdingqi/home/foursquare-dataset},
  visted on May 25, 2022.

\bibitem{brightkite}
``Brightkite check-in datasets,''
  \url{https://snap.stanford.edu/data/loc-brightkite.html}, visted on May 25,
  2022.

\bibitem{gowalla}
``Gowalla check-in datasets,''
  \url{https://snap.stanford.edu/data/loc-Gowalla.html}, visted on May 25,
  2022.

\bibitem{singaporedata}
``Data sets of land transport authority,''
  \url{https://datamall.lta.gov.sg/content/datamall/en.html}, visted on May 25,
  2022.

\bibitem{traveltimedata}
``Travel time, speed, and mobility heatmap data provided by uber,''
  \url{https://movement.uber.com/explore/amsterdam/travel-times/query}, visted
  on May 25, 2022.

\bibitem{didisimulation}
``Didi open-sourced simulation platform,''
  \url{https://outreach.didichuxing.com/Simulation/}, visted on May 25, 2022.

\bibitem{ruch2018amodeus}
C.~Ruch, S.~H{\"o}rl, and E.~Frazzoli, ``Amodeus, a simulation-based testbed
  for autonomous mobility-on-demand systems,'' in \emph{Proc.~of IEEE ITSC},
  2018.

\bibitem{ruch2020quantifying}
C.~Ruch, C.~Lu, L.~Sieber, and E.~Frazzoli, ``Quantifying the efficiency of
  ride sharing,'' \emph{IEEE Transactions on Intelligent Transportation
  Systems}, vol.~22, no.~9, pp. 5811--5816, 2021.

\bibitem{amod2}
``A high-capacity on-demand ride-sharing simulator, with three representative
  vehicle dispatch algorithms implemented,''
  https://github.com/sustech-isus/AMoD2, visted on May 25, 2022.

\bibitem{li2021optimal}
C.~Li, D.~Parker, and Q.~Hao, ``Optimal online dispatch for high-capacity
  shared autonomous mobility-on-demand systems,'' in \emph{Proc.~of IEEE ICRA},
  2021.

\bibitem{abm-1.0}
``Amod-abm, an agent-based modeling platform for simulating autonomous
  mobility-on-demand systems,'' \url{https://github.com/wenjian0202/amod-abm}.

\bibitem{abm-2.0}
``Mod-abm-2.0, an agent-based modeling platform for mobility-on-demand
  simulations,'' \url{https://https://github.com/wenjian0202/mod-abm-2.0},
  visted on May 25, 2022.

\bibitem{wen2017rebalancing}
J.~Wen, J.~Zhao, and P.~Jaillet, ``Rebalancing shared mobility-on-demand
  systems: A reinforcement learning approach,'' in \emph{Proc.~of IEEE ITSC},
  2017.

\bibitem{horni2016multi}
A.~Horni, K.~Nagel, and K.~W. Axhausen, \emph{The multi-agent transport
  simulation MATSim}.\hskip 1em plus 0.5em minus 0.4em\relax Ubiquity Press,
  2016.

\bibitem{tsao2019model}
M.~Tsao, D.~Milojevic, C.~Ruch, M.~Salazar, E.~Frazzoli, and M.~Pavone, ``Model
  predictive control of ride-sharing autonomous mobility-on-demand systems,''
  in \emph{Proc.~of IEEE ICRA}, 2019.

\bibitem{lopez2018microscopic}
P.~A. Lopez, M.~Behrisch, L.~Bieker-Walz, J.~Erdmann, Y.-P. Fl{\"o}tter{\"o}d,
  R.~Hilbrich, L.~L{\"u}cken, J.~Rummel, P.~Wagner, and E.~Wie{\ss}ner,
  ``Microscopic traffic simulation using sumo,'' in \emph{Proc.~of IEEE ITSC},
  2018.

\bibitem{castagna2021multi}
A.~Castagna and I.~Dusparic, ``Multi-agent transfer learning in reinforcement
  learning-based ride-sharing systems,'' in \emph{Proc.~of ICAART}, 2022.

\bibitem{zhu2021shared}
L.~Zhu, Z.~Zhao, and G.~Wu, ``Shared automated mobility with demand-side
  cooperation: A proof-of-concept microsimulation study,''
  \emph{Sustainability}, vol.~13, no.~5, p. 2483, 2021.

\bibitem{zhang2019cityflow}
H.~Zhang, S.~Feng, C.~Liu, Y.~Ding, Y.~Zhu, Z.~Zhou, W.~Zhang, Y.~Yu, H.~Jin,
  and Z.~Li, ``Cityflow: A multi-agent reinforcement learning environment for
  large scale city traffic scenario,'' in \emph{Proc.~of WWW}, 2019.

\bibitem{ota2016stars}
M.~Ota, H.~Vo, C.~Silva, and J.~Freire, ``Stars: Simulating taxi ride sharing
  at scale,'' \emph{IEEE Transactions on Big Data}, vol.~3, no.~3, pp.
  349--361, 2016.

\bibitem{mounesan2021fleet}
M.~Mounesan, V.~Jayawardana, Y.~Wu, S.~Samaranayake, and H.~T. Vo, ``Fleet
  management for ride-pooling with meeting points at scale: a case study in the
  five boroughs of new york city,'' \emph{arXiv preprint arXiv:2105.00994},
  2021.

\bibitem{khalil2022realistic}
J.~Khalil, D.~Yan, L.~Yuan, M.~Jafarzadehfadaki, S.~Adhikari, V.~P. Sisiopiku,
  and Z.~Jiang, ``Realistic urban traffic simulation with ride-hailing
  services: a revisit to network kernel density estimation (systems paper),''
  in \emph{Proc.~of ACM SIGSPATIAL}, 2022.

\bibitem{salman2023quantifying}
F.~Salman, V.~P. Sisiopiku, J.~Khalil, M.~Jafarzadehfadaki, and D.~Yan,
  ``Quantifying the impact of transportation network companies on urban
  congestion in a medium sized city,'' \emph{Journal of Traffic and
  Transportation Engineering}, vol.~11, pp. 1--14, 2023.

\bibitem{openai-gym}
G.~Brockman, V.~Cheung, L.~Pettersson, J.~Schneider, J.~Schulman, J.~Tang, and
  W.~Zaremba, ``Openai gym,'' \emph{arXiv preprint arXiv:1606.01540}, 2016.

\bibitem{liu2022smart}
S.~Liu, Y.~Wang, X.~Chen, Y.~Fu, and X.~Di, ``Smart-eflo: An integrated
  sumo-gym framework for multi-agent reinforcement learning in electric fleet
  management problem,'' in \emph{Proc.~of IEEE ITSC}, 2022.

\bibitem{openstreetmap}
``Openstreetmap,'' \url{https://www.openstreetmap.org/}, visted on May 25,
  2022.

\bibitem{boeing2017osmnx}
G.~Boeing, ``Osmnx: New methods for acquiring, constructing, analyzing, and
  visualizing complex street networks,'' \emph{Computers, Environment and Urban
  Systems}, vol.~65, pp. 126--139, 2017.

\bibitem{weatherdata}
``Data tools: Local climatological data ({LCD}),''
  \url{https://www.ncdc.noaa.gov/cdo-web/datatools/lcd}, visted on May 25,
  2022.

\bibitem{kddcup20}
``Learning to dispatch and reposition on a mobility-on-demand platform,''
  \url{https://www.biendata.xyz/competition/kdd_didi/}, visted on May 25, 2022.

\bibitem{carron2019scalable}
A.~Carron, F.~Seccamonte, C.~Ruch, E.~Frazzoli, and M.~N. Zeilinger, ``Scalable
  model predictive control for autonomous mobility-on-demand systems,''
  \emph{IEEE Transactions on Control Systems Technology}, 2019.

\bibitem{bischoff2016simulation}
J.~Bischoff and M.~Maciejewski, ``Simulation of city-wide replacement of
  private cars with autonomous taxis in berlin,'' \emph{Procedia computer
  science}, vol.~83, pp. 237--244, 2016.

\bibitem{allan2015benchmark}
D.~F. Allan and A.~M. Farid, ``A benchmark analysis of open source
  transportation-electrification simulation tools,'' in \emph{Proc.~of IEEE
  ITSC}, 2015.

\bibitem{mao2020dispatch}
C.~Mao, Y.~Liu, and Z.-J.~M. Shen, ``Dispatch of autonomous vehicles for taxi
  services: A deep reinforcement learning approach,'' \emph{Transportation
  Research Part C: Emerging Technologies}, vol. 115, p. 102626, 2020.

\bibitem{mo2022modeling}
D.~Mo, X.~M. Chen, and J.~Zhang, ``Modeling and managing mixed on-demand ride
  services of human-driven vehicles and autonomous vehicles,''
  \emph{Transportation Research Part B: Methodological}, vol. 157, pp. 80--119,
  2022.

\bibitem{fan2022joint}
G.~Fan, H.~Jin, Y.~Zhao, Y.~Song, X.~Gan, J.~Ding, L.~Su, and X.~Wang, ``Joint
  order dispatch and charging for electric self-driving taxi systems,'' in
  \emph{Proc.~of IEEE INFOCOM}, 2022.

\bibitem{zhang2022autonomous}
S.~Zhang, C.~Markos, and J.~James, ``Autonomous vehicle intelligent system:
  joint ride-sharing and parcel delivery strategy,'' \emph{IEEE Transactions on
  Intelligent Transportation Systems}, vol.~23, no.~10, pp. 18\,466--18\,477,
  2022.

\bibitem{cnn2014uber}
``Uber's dirty tricks quantified: Rival counts 5,560 canceled rides,''
  \url{https://money.cnn.com/2014/08/11/technology/uber-fake-ride-requests-lyft/index.html},
  visted on May 25, 2022.

\bibitem{zhao2020blackbox}
Y.~Zhao, I.~Shumailov, H.~Cui, X.~Gao, R.~Mullins, and R.~Anderson, ``Blackbox
  attacks on reinforcement learning agents using approximated temporal
  information,'' in \emph{Proc.~of IEEE/IFIP DSN-W}, 2020.

\bibitem{cook2018gender}
C.~Cook, R.~Diamond, J.~Hall, J.~A. List, and P.~Oyer, ``The gender earnings
  gap in the gig economy: Evidence from over a million rideshare drivers,''
  National Bureau of Economic Research, Tech. Rep., 2018.

\bibitem{tjaden2018ride}
J.~D. Tjaden, C.~Schwemmer, and M.~Khadjavi, ``Ride with me—ethnic
  discrimination, social markets, and the sharing economy,'' \emph{European
  Sociological Review}, vol.~34, no.~4, pp. 418--432, 2018.

\bibitem{wang2020disruptive}
S.~Wang and M.~Smart, ``The disruptive effect of ridesourcing services on
  for-hire vehicle drivers’ income and employment,'' \emph{Transport Policy},
  vol.~89, pp. 13--23, 2020.

\bibitem{wolfson2017fairness}
O.~Wolfson and J.~Lin, ``Fairness versus optimality in ridesharing,'' in
  \emph{Proc.~of IEEE MDM}, 2017.

\bibitem{lesmana2019balancing}
N.~Lesmana, X.~Zhang, and X.~Bei, ``Balancing efficiency and fairness in
  on-demand ridesourcing,'' in \emph{Proc.~of NeurIPS}, 2019.

\bibitem{xu2020trade}
Y.~Xu and P.~Xu, ``Trade the system efficiency for the income equality of
  drivers in rideshare,'' in \emph{Proc.~of IJCAI}, 2020.

\bibitem{nanda2020balancing}
V.~Nanda, P.~Xu, K.~A. Sankararaman, J.~Dickerson, and A.~Srinivasan,
  ``Balancing the tradeoff between profit and fairness in rideshare platforms
  during high-demand hours,'' in \emph{Proc.~of AAAI}, 2020.

\bibitem{suhr2019two}
T.~S{\"u}hr, A.~J. Biega, M.~Zehlike, K.~P. Gummadi, and A.~Chakraborty,
  ``Two-sided fairness for repeated matchings in two-sided markets: A case
  study of a ride-hailing platform,'' in \emph{Proc.~of ACM SIGKDD}, 2019.

\bibitem{raman2021data}
N.~Raman, S.~Shah, and J.~P. Dickerson, ``Data-driven methods for balancing
  fairness and efficiency in ride-pooling,'' in \emph{Proc.~of IJCAI}, 2021.

\bibitem{barocas2017fairness}
S.~Barocas, M.~Hardt, and A.~Narayanan, ``Fairness in machine learning,''
  \emph{Nips tutorial}, vol.~1, p.~2, 2017.

\bibitem{zhao2019preference}
B.~Zhao, P.~Xu, Y.~Shi, Y.~Tong, Z.~Zhou, and Y.~Zeng, ``Preference-aware task
  assignment in on-demand taxi dispatching: An online stable matching
  approach,'' in \emph{Proc.~of AAAI}, 2019.

\bibitem{xu2020unified}
Y.~Xu, P.~Xu, J.~Pan, and J.~Tao, ``A unified model for the two-stage
  offline-then-online resource allocation,'' in \emph{Proc.~of IJCAI}, 2020.

\bibitem{lowalekar2020competitive}
M.~Lowalekar, P.~Varakantham, and P.~Jaillet, ``Competitive ratios for online
  multi-capacity ridesharing,'' in \emph{Proc.~of AAMAS}, 2020.

\bibitem{lykouris2018competitive}
T.~Lykouris and S.~Vassilvtiskii, ``Competitive caching with machine learned
  advice,'' in \emph{Proc.~of ICML}, 2018.

\bibitem{purohit2018improving}
M.~Purohit, Z.~Svitkina, and R.~Kumar, ``Improving online algorithms via ml
  predictions,'' in \emph{Proc.~of NeurIPS}, 2018.

\bibitem{anand2020customizing}
K.~Anand, R.~Ge, and D.~Panigrahi, ``Customizing ml predictions for online
  algorithms,'' in \emph{Proc.~of ICML}, 2020.

\bibitem{NEURIPS2020_e834cb11}
E.~Bamas, A.~Maggiori, and O.~Svensson, ``The primal-dual method for learning
  augmented algorithms,'' in \emph{Proc.~of NeurIPS}, 2020.

\bibitem{rohatgi2020near}
D.~Rohatgi, ``Near-optimal bounds for online caching with machine learned
  advice,'' in \emph{Proc.~of ACM-SIAM SODA}, 2020.

\bibitem{mitzenmacher2020algorithms}
M.~Mitzenmacher and S.~Vassilvitskii, ``Algorithms with predictions,'' in
  \emph{Beyond the Worst-Case Analysis of Algorithms}, T.~Roughgarden,
  Ed.\hskip 1em plus 0.5em minus 0.4em\relax Cambridge University Press, 2020,
  ch.~30.

\bibitem{wu2020comprehensive}
Z.~Wu, S.~Pan, F.~Chen, G.~Long, C.~Zhang, and S.~Y. Philip, ``A comprehensive
  survey on graph neural networks,'' \emph{IEEE transactions on neural networks
  and learning systems}, 2020.

\bibitem{li2018diffusion}
Y.~Li, R.~Yu, C.~Shahabi, and Y.~Liu, ``Diffusion convolutional recurrent
  neural network: Data-driven traffic forecasting,'' in \emph{Proc.~of ICLR},
  2018.

\bibitem{yu2018spatio}
B.~Yu, H.~Yin, and Z.~Zhu, ``Spatio-temporal graph convolutional networks: A
  deep learning framework for traffic forecasting,'' in \emph{Proc.~of IJCAI},
  2017.

\bibitem{yao2018deep}
H.~Yao, F.~Wu, J.~Ke, X.~Tang, Y.~Jia, S.~Lu, P.~Gong, J.~Ye, and Z.~Li, ``Deep
  multi-view spatial-temporal network for taxi demand prediction,'' in
  \emph{Proc.~of AAAI}, 2018.

\bibitem{kim2020idle}
S.~Kim, U.~Lee, I.~Lee, and N.~Kang, ``Idle vehicle relocation strategy through
  deep learning for shared autonomous electric vehicle system optimization,''
  \emph{Journal of Cleaner Production}, vol. 333, p. 130055, 2022.

\bibitem{li2022decentralized}
B.~Li, N.~Ammar, P.~Tiwari, and H.~Peng, ``Decentralized ride-sharing of shared
  autonomous vehicles using graph neural network-based reinforcement
  learning,'' in \emph{Proc.~of IEEE ICRA}, 2022.

\end{thebibliography}

\vspace{-0.4cm}

\begin{IEEEbiography}[{\includegraphics[width=1in,height=1.25in,clip,keepaspectratio]{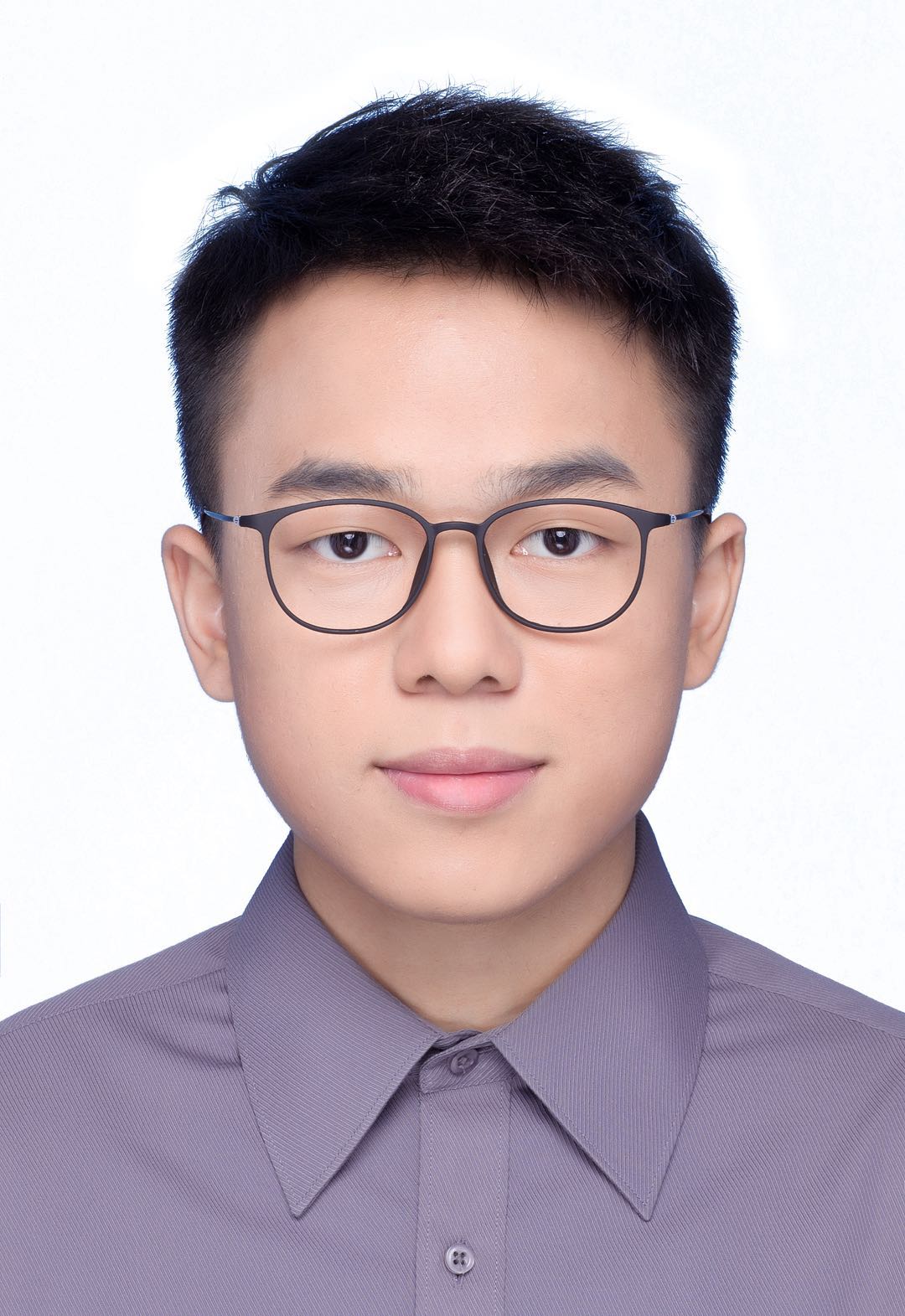}}]{Dacheng Wen}
(Student Member, IEEE) received the M.S.~degree in computer science from The University of Hong Kong. 
He is currently pursing the Ph.D.~degree with the Department of Computer Science, The University of Hong Kong.
His research interests include machine learning and intelligent transportation systems.
He is also excited about the emerging blockchain-driven revolution of various computer science and information technology.
\end{IEEEbiography}

\vspace{-0.4cm}

\begin{IEEEbiography}[{\includegraphics[width=1in,height=1.25in,clip,keepaspectratio]{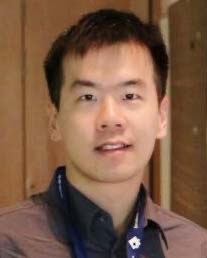}}]{Yupeng Li} (Member, IEEE) received the Ph.D. degree in computer science from The University of Hong Kong. He was with the University of Toronto and is currently with Hong Kong Baptist University. His research interests are in general areas of network science and, in particular, algorithmic decision making and machine learning problems, which arise in networked systems. 
He is also excited about interdisciplinary research that applies algorithmic techniques to edging problems. Recently, he has worked on robust online machine learning for the application of data classification, and he has extended these techniques to modern areas in networking and social media. Dr. Li has been awarded the Rising Star in Social Computing Award by CAAI and the distinction of Distinguished Member of the IEEE INFOCOM Technical Program Committee in 2022. He serves on the technical committees of some top conferences in computer science. His works have been published in prestigious venues, such as \textsc{IEEE INFOCOM}, \textsc{ACM MobiHoc}, \textsc{IEEE Journal on Selected Areas in Communications}, and \textsc{IEEE/ACM Transactions on Networking}. He is a member of ACM and IEEE.
\end{IEEEbiography}

\vspace{-0.4cm}

\begin{IEEEbiography}[{\includegraphics[width=1in,height=1.25in,clip,keepaspectratio]{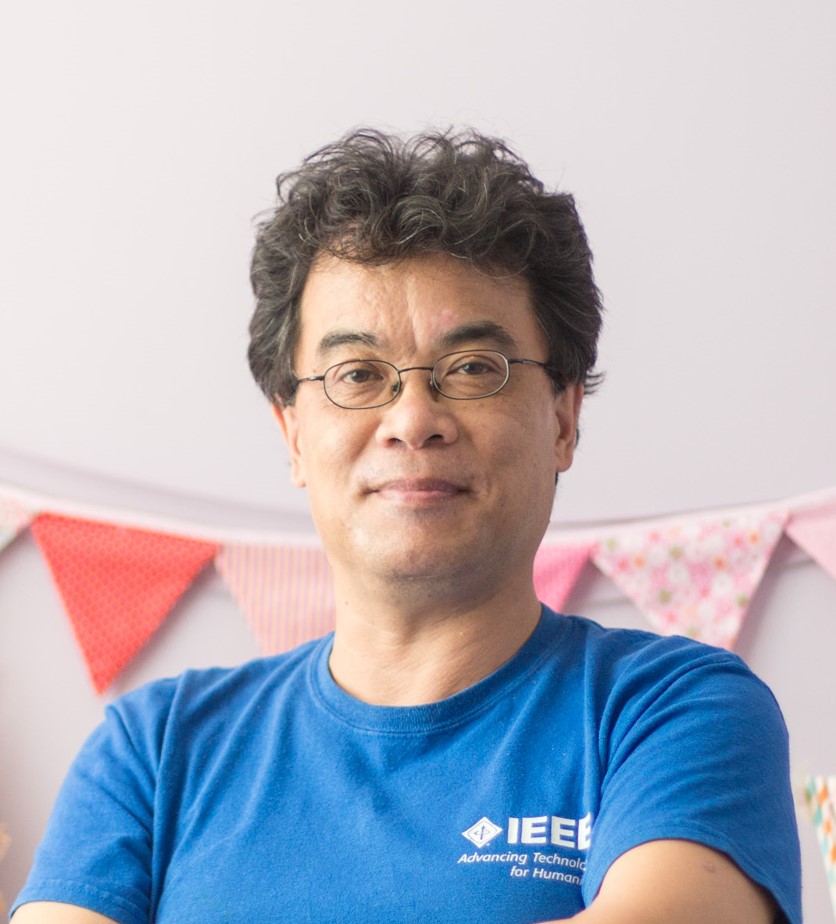}}]{Francis C.M. Lau} received the Ph.D. degree from the Department of Computer Science, University of Waterloo. He is currently an Honorary Professor and Associate Head (part-time) of the Department of Computer Science at The University of Hong Kong, China. His research interests include computer systems, networks, programming languages, and application of computing in arts. He was the Editor-in-Chief of the Journal of Interconnection Networks from 2011 to 2020. 
\end{IEEEbiography}
 
\end{document}